\documentclass[runningheads]{llncs}

\usepackage{eccv}

\usepackage{eccvabbrv}

\usepackage{graphicx}
\usepackage{booktabs}

\usepackage[accsupp]{axessibility}  

\usepackage{hyperref}

\usepackage{orcidlink}

\begin{document}

\title{	
Free-Lunch Augmentation by Revisiting Diffusion-Based Data Generation for Cross-Domain Few-Shot Object Detection} 

\titlerunning{Revisiting Diffusion-Based Data Generation for CDFSOD}


\author{Zijian Zhuang\inst{\dagger} \and
Yixiong Zou\inst{\dagger}\thanks{Corresponding author, $\dagger~$Equal contribution} \and 
Yuhua Li \and
Ruixuan Li}

\authorrunning{Z. Zhuang, Y. Zou, et al.}

\institute{School of Computer Science and Technology, Huazhong University of Science and Technology, China \\
\email{\{zhuangzijian, yixiongz, idcliyuhua, rxli\}@hust.edu.cn}\\}

\maketitle

\begin{abstract}
Cross-Domain Few-Shot Object Detection (CDFSOD) aims to transfer knowledge from data-rich upstream generic domains to downstream expert domains using scarce training data, where the significant domain gap and data scarcity make it an unsolved challenge.
To address this problem, we revisit a natural yet underexplored approach in CDFSOD: data augmentation, by directly synthesizing data through diffusion models to supplement limited training samples.
However, due to large domain gaps, we find that current diffusion methods cannot produce good results, leading to performance even lower than using the original images.
To address these limitations, we divide the domain gaps into visual gaps and semantic gaps for separate analysis.
For the visual gap, we find that the diffusion model cannot distinguish noise from useful information on expert domains, which can be mitigated by adding weakened noise.
For the semantic gap, we find that the background semantics shows much smaller gaps between domains than foreground semantics, and we can bridge this gap by background inpainting.
Based on the above analysis, we propose a method (Selective Inpainting with Tailored Noise, SITN) to dynamically take different strategies for downstream data synthesis based on their different gaps from the general domain, including a Generation Module for adding tailored noise and a Selection Module to dynamically select the inpainting regions.
Extensive experiments on 6 datasets of CDFSOD and 4 datasets of cross-domain few-shot segmentation (CDFSS) validate that we can synthesize helpful data, achieving new state-of-the-art performance.
Our codes is available at \url{https://github.com/zzzzj311-droid/Free-Lunch-SITN}.
\end{abstract}

\section{Introduction}
\label{sec:intro}
\vspace{-0.3cm}

In real-world scenarios, obtaining sufficient data is often challenging for downstream expert-domain tasks~\cite{zou2024flatten}\cite{yan2026start}\cite{zou2024compositional}, such as medical analysis~\cite{Zhao_2026_CVPR}~\cite{huang2024sparse}~\cite{zou2024attention}. To handle this problem, Cross-Domain Few-Shot Object Detection (CDFSOD)~\cite{xiong2023cd}\cite{jiang2026remedying} has been proposed to transfer models pretrained on large-scale upstream datasets to downstream data-scarce expert datasets for object detection, where the main challenge lies in both the downstream data scarcity and domain gaps between downstream and upstream datasets.

\begin{figure}[t]
    \centering
    \includegraphics[width=1\linewidth]{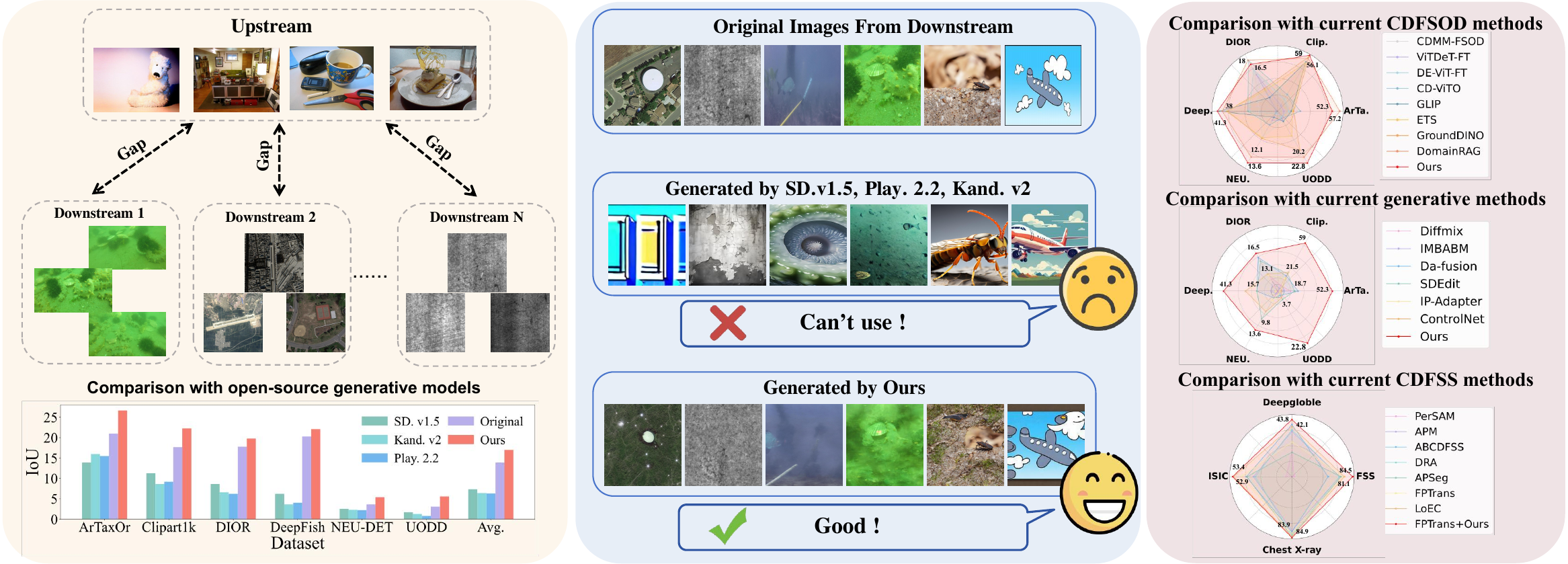}
    \caption{(Left-top) Cross-Domain Few-Shot Object Detection learns from scarce training data on downstream expert domains, with knowledge generalized from upstream general domains, where domain gaps and data scarcity make it challenging. 
        (Left-bottom \& Mid) To address it, we revisit a natural but ignored approach in CDFSOD (i.e., data augmentation) to directly supplement the scarce training data with diffusion models. 
        However, due to large domain gaps, we find that current methods can hardly synthesize satisfactory samples, leading to a performance even lower than merely using the original data, which we aim to address in this paper.
        (Right) While existing diffusion-based methods are not directly applicable to CDFSOD, our method not only addresses this task effectively but also seamlessly extends to CDFSS. It achieves new state-of-the-art results on both benchmarks, demonstrating its versatility and superiority.        
        }
        \vspace{-0.5cm}
    \label{fig: placeholder}
\end{figure}

To handle this problem, current state-of-the-art methods primarily focus on transfer learning or efficient fine-tuning. However, driven by recent progress in AIGC~\cite{foo2025ai}, a straightforward method for directly addressing the data scarcity is still under-explored: \textbf{directly synthesizing the scarce training data}. However, it also faces great challenges. As in Fig.~\ref{fig: placeholder}, we find that directly applying standard image-to-image or text-to-image generation cannot produce suitable images due to large domain gaps. Take Stable Diffusion v1.5~\cite{rombach2022high} for example, although its training data covers a wide range of images, it still cannot contain sufficient downstream expert-domain data, such as medical data, which are expensive to obtain and may not be open to society due to privacy issues. 
Therefore, the generated data proves unsatisfactory, thereby degrading the performance.

In this paper, we aim to address this data synthesis problem in diffusion models caused by huge domain gaps. We divide the domain gap into the visual gap (e.g., general images vs. X-Ray images) and the semantic gap (e.g., COCO classes vs. medical classes), and analyze them separately.

For the visual gap, given expert-domain datasets, we find that the difference between synthesized data and the original data is much larger than that on the general-domain datasets, which we explain as the model's inability to distinguish noise and useful information on expert domains. Then, we find that adding just weak noises can effectively maintain the useful information in the given expert-domain images, mitigating the negative effect of visual gaps.

For the semantic gap, as most categories do not appear in the diffusion model, the model can hardly comprehend the unknown semantics. However, we then find these categories are mostly in the foreground of the image, but the background is more likely to be shared across domains. For example, a general domain may contain a dog on a grassland, and a target domain may contain an alien against a green background. Indeed, the foreground object (dog vs. alien) shows large semantic gaps, but their backgrounds (grassland vs. green background) are more likely to share similar patterns. Therefore, it is possible to utilize background inpainting to circumvent the synthesis of unknown semantics.

Based on this analysis, we propose our method (Selective Inpainting with Tailored Noise, SITN) to tailor current pretrained diffusion models for data synthesis of expert-domain data. Specifically, our method contains a Generation Module, which synthesizes images with noises tailored (e.g., weakened) for mitigating the negative effect of domain gaps, and a Selection Module, which dynamically selects whether the background inpainting or the foreground inpainting strategy should be adopted. As shown in Fig.~\ref{fig: placeholder}, our method successfully supplements data for the CDFSOD task and achieves new state-of-the-art performance on six benchmark datasets. Our contribution can be listed as 
\begin{itemize}
    \item To the best of our knowledge, we are the first to study the data synthesis problem under large domain gaps for CDFSOD. \textit{Experiments show that our conclusion also fits the cross-domain few-shot segmentation (CDFSS) task}.
    \item By extensive experiments, we propose to take weakened noises to reduce the negative effect of visual gaps, and utilize background inpainting to handle semantic gaps.
    \item We propose a method (Selective Inpainting with Tailored Noise, SITN) to dynamically take different strategies for downstream data synthesis based on their different gaps from the general domain.
    \item Extensive experiments validate that we can successfully synthesize helpful data for the CDFSOD task and even the CDFSS task, achieving new state-of-the-art methods on 6 CDFSOD datasets and 4 CDFSS datasets.
\end{itemize}

\section{Related Work}
\vspace{-0.3cm}
\textbf{Cross-Domain Few-Shot Object Detection (CDFSOD)} aims to effectively learn the target domain using only a small amount of training samples~\cite{fu2024cross}, where the domain gap and scarce training data make it challenging.
\cite{bou2024exploring} proposes a prototype-based few-shot object detection method specifically designed to address the scarcity of data in satellite imagery. \cite{gui2024few} proposes the UPPR method to address the problems of base class forgetting and poor novel class detection in Generalized Few-Shot Object Detection.
FM-FSOD~\cite{han2024few} integrates the DINOv2 visual backbone with the contextual learning capabilities of large language models.
However, data synthesis is still under-explored in CDFSOD.

\vspace{0.1cm}
\noindent\textbf{Diffusion Models} are to learn the distribution of the training dataset via denoising, thereby generating images of the same distribution~\cite {cao2024survey}. Current diffusion-based methods relevant to our task focus on source-domain training for transfer learning~\cite {jabbour2025depict}, Multi-sample semantic segmentation~\cite {shen2024cgmgm}, and Multi-sample image inpainting~\cite {fei2023generative}. 
For generation fields, most methods are based on image-to-image generation~\cite {liu2024residual,gui2024few} with whole-noised information, text-to-image generation~\cite {go2023towards,zhang2023prompt,sain2023clip}.
However, applying diffusion models to handle the data scarcity problem for CDFSOD is still underexplored, and our method is the first to hold the cross-domain few-shot data synthesis under large domain gaps. 

\vspace{0.1cm}
\noindent\textbf{Data Augmentation} is widely adopted by current works.
Recent advances in diffusion models have significantly enhanced data augmentation techniques for image generation and semantic editing. DA-Fusion~\cite{trabucco2023effective} employs a pre-trained text-to-image diffusion model to perform semantic editing on original images.
IMBABM~\cite{he2022synthetic} systematically investigates the applicability of synthetic data generated by text-to-image models in image recognition tasks.
DIFFUSEMIX~\cite{islam2024diffusemix} employs diffusion models to create augmented images through generation, splicing, and fractal mixing processes.
IP-Adapter~\cite{ye2023ip} is a lightweight adapter that adds image prompt capabilities to pre-trained text-to-image diffusion models. ControlNET~\cite{zhang2023adding} locks the parameters of the original model and trains a trainable copy of its encoding layers. 
\noindent\textbf{SDEdit}~\cite{meng2021sdedit} formulates the image editing task via a stochastic differential equation (SDE) to simulate the reverse-time stochastic process. However, its generation process is inherently constrained by the training data distribution, struggling to generalize to cross-domain tasks (e.g., editing real-world photos using medical image priors), and its backbone network is limited in scale. Even if retrained on cross-domain datasets, the stochastic gradient descent optimization within the SDE paradigm remains computationally prohibitive. In contrast, our method is training-free and plug-and-play. It is not only compatible with existing open-source generative priors to mitigate domain gaps but also significantly outperforms SDEdit in inference efficiency, overcoming the dual limitations of SDE-based methods in domain generalization and computational cost.
\noindent\textbf{DomainRAG}~\cite{li2025domain} adopts a retrieval-augmented pipeline that heavily relies on large-scale source-domain datasets such as COCO, requiring background image screening and ResNet-50 feature extraction, which significantly increases inference time. In contrast, our method is based on the purely generative DDIM paradigm, eliminating the need for external retrieval and feature matching, thereby achieving substantially higher generation efficiency and making it more suitable for real-time or large-scale applications.

\section{Why diffusion models fail in cross-domains?}
\vspace{-0.3cm}

\subsection{Negative Effect of Domain Gaps}
\vspace{-0.2cm}

\begin{figure}[htbp]
    \centering
    \includegraphics[width=1.0\linewidth]{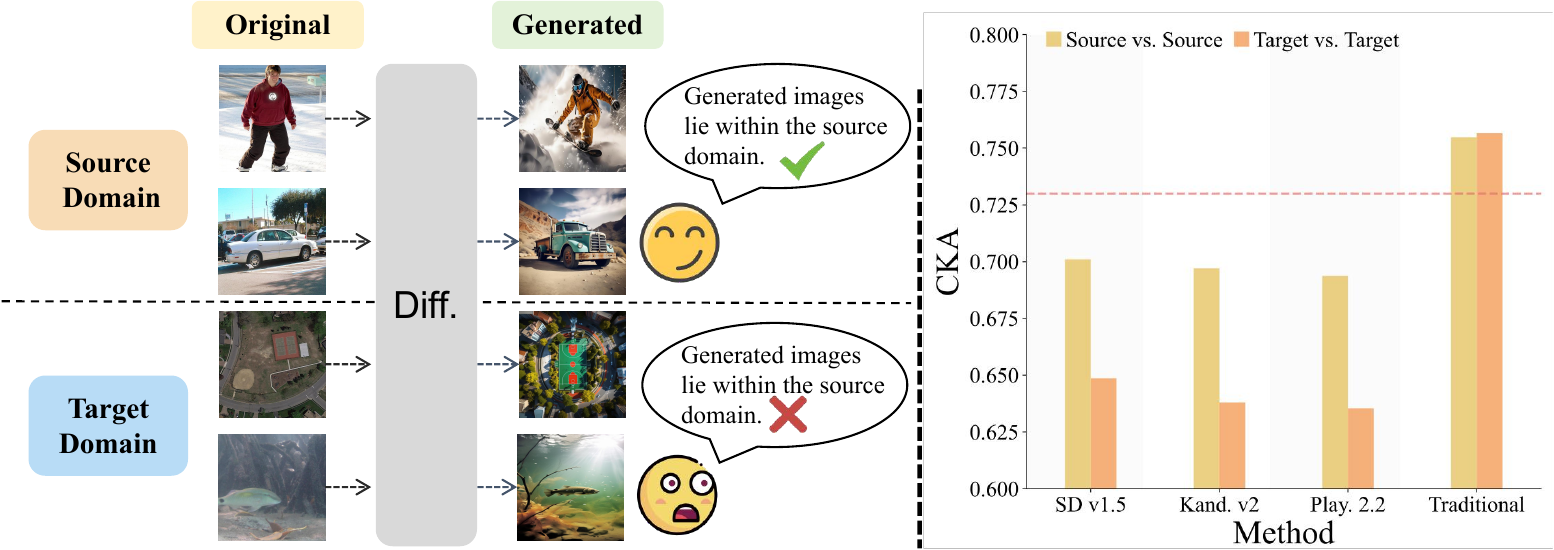}
    \caption{(Left) Synthesized images on source domains well preserve the semantics of the original images, but largely destroy the semantics on the target domains.   
    (Right) To quantify such a difference between domains, we use the CKA similarity to measure the semantic similarity between synthesized and original images. We can see the traditional method (e.g., flip, rotation) shows consistent CKA similarities across domains, verifying its robustness to domains, but diffusion-based ones suffer from a large decline in target domains, showing the difficulty brought by domain gaps.}
    \label{fig: Source vs Target}
    \vspace{-0.5cm}
\end{figure}

Fig.~\ref{fig: placeholder} shows that images synthesized by current methods are largely different from the original ones. 
To delve into this phenomenon, we first plot images synthesized on source domains and target domains in Fig.~\ref{fig: Source vs Target} (Left). We can see that the difference between the synthesized and original images is much smaller than that in the target domains.

To quantify this difference, we follow~\cite{Davari_Horoi_Natik_Lajoie_Wolf_Belilovsky_2022} to take the CKA (Centered Kernel Alignment) similarity to measure the similarity between synthesized and original images. We use a vision encoder to extract features from images and calculate the CKA similarity between them.
Here, we compare mainstream generative models (including Stable-Diffusion v1.5, Playground v2, and Kandinsky 2.2) and the traditional image augmentation methods (i.e., horizontal/vertical flipping). As in Fig.~\ref{fig: Source vs Target}  (Right), for the traditional augmentation methods, there is a consistency between the source and target domains, which naturally serves as an ``oracle" for other synthesis methods due to its wide adoption\footnote{There is a trade-off between maintaining original information and introducing augmented information. Traditional methods are effective at the former but less effective at the latter; therefore are not the optimal choice.}.

However, for other diffusion-based models, there exists a huge gap between the CKA on source domains and target domains, indicating the challenges in data generation. 
That is, we hypothesize that the model can hardly distinguish noise from useful visual information in target-domain images; therefore, it just ``randomly"  guesses what to synthesize based on its pre-training information. However, the pre-training information also largely differs from the target-domain information. As a result, it cannot successfully supplement the scarce training data in Fig.~\ref{fig: placeholder}.

In the following subsections, we will divide the domain gap into the visual gap (e.g., natural images vs. underwater images) and the semantic gap (e.g., ImageNet classes vs. medical classes), and analyze them separately.

\vspace{-0.3cm}
\subsection{Visual Gaps}
\vspace{-0.3cm}

To verify our hypothesis about the noisy information, we then try to control the strength of noise added to the diffusion model and evaluate the CKA similarity between the synthesized and original images.

Specifically, diffusion models generate images in two stages. In the forward stage, noise is gradually added to the image, progressively obscuring the information in the image. In the backward stage, the diffusion model gradually removes the noise and utilizes the knowledge learned from the source domain to infer the obscured information. Therefore, we can control the noise added in the forward stage and use the backward stage to generate images. 
\begin{figure*}[t]
    \centering
    
    \includegraphics[width=1.0\linewidth]{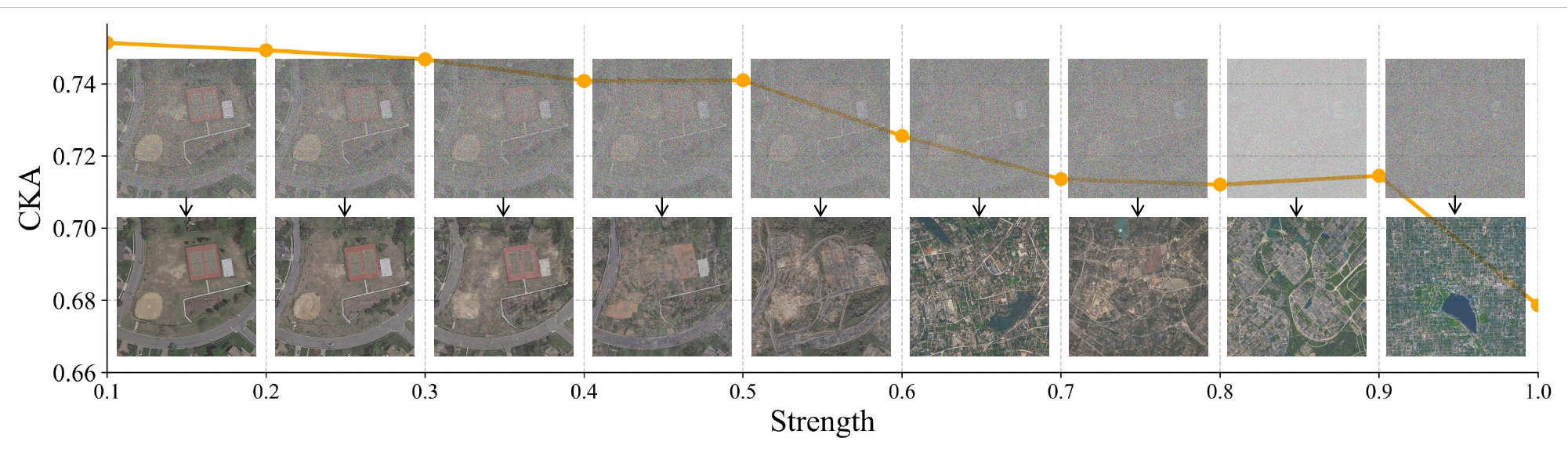}
    \vspace{-0.6cm}
    \caption{The relationship between noise strength and CKA similarity shows that low noise strength effectively preserves the semantics of the original image (i.e., similarity only slightly decreases), while high noise levels severely disrupt semantics (i.e., generated images differ greatly from the original one), indicating diffusion models can hardly recover target-domain information from the noise.}
    
    \label{fig: Strength vs. CKA}
    \vspace{-0.5cm}
    
\end{figure*}

Fig.~\ref{fig: Strength vs. CKA} shows the generated images and the corresponding CKA. By gradually increasing the strength of noise, the generated images gradually deviate from the original images, with more information synthesized by the diffusion model. Simultaneously, the CKA value consistently decreases.
In contrast, by reducing the strength of noise, more semantic information is preserved, and the CKA gradually increases towards that of the traditional augmentation methods in Fig.~\ref{fig: Strength vs. CKA}.
By controlling noise strength, we could directly guide the diffusion model to preserve content and bridge visual gaps, verifying our hypothesis.

\vspace{-0.3cm}

\subsection{Semantic Gaps}
\vspace{-0.3cm}

For the semantic gap, since the target-domain category may not appear in the diffusion model's training data, it is difficult for the diffusion model to synthesize such an unknown category. However, we find that an intuition naturally holds that the category information is mostly relevant to the foreground object, but for the background part, it is easier to be shared across domains.
For example, a general domain may contain a dog on a grassland, and a target domain may contain an alien against a green background. Indeed, the foreground object (dog vs. alien) shows large semantic gaps, but their backgrounds (grassland vs. green background) are more likely to share similar patterns. Therefore, it is possible to utilize background inpainting to circumvent the synthesis of unknown semantics.

\begin{figure}[t]
    \centering
    \includegraphics[width=1.0\linewidth]{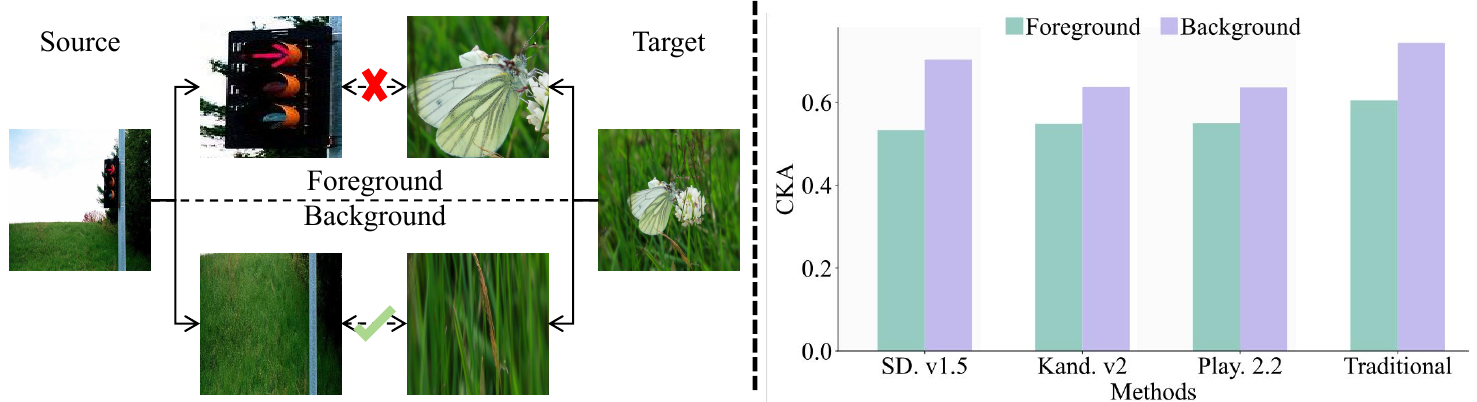}
    \caption{ (Left) Compared with the foreground relevant to unseen classes, the background is easier to transfer across domains.
        (Right) To verify this intuition, we crop the foreground and background for data synthesis and find that the background CKA is consistently higher than that of the foreground, indicating it is possible to circumvent the semantic gap by background inpainting.}
    \label{fig: Semantic Gaps}
    \vspace{-0.5cm}
    
\end{figure}

To verify this intuition, in Fig.~\ref{fig: Semantic Gaps} (Left), we try to use the bounding box annotation to crop the foreground and background parts in the target-domain images, and use diffusion models to synthesize the foreground and background images. Then, we also evaluate the CKA similarity between the synthesized and original images. We can see that for different diffusion models, the background CKA is consistently much higher than that of the foreground, verifying our intuition that the background inpainting could be much easier on the target domains, even when the semantic gap is large.

\section{Method}
\vspace{-0.3cm}
Based on the above analysis, we can see that both the visual gap and the semantic gap make diffusion models fail to synthesize for the CDFSOD task. Moreover, we find that controlling the noise strength in diffusion models can effectively reduce the visual gap, while employing background generation can minimize the semantic gap between synthesized images and original ones. Therefore, we further propose a novel image generation method capable of producing images that closely resemble the original CDFSOD dataset in both visual and semantic features, which effectively supplements training data for the CDFSOD task.
\vspace{-0.2cm}

\vspace{-0.3cm}

\subsection{Problem Definition}
\vspace{-0.2cm}

\subsubsection{Cross-Domain Few-Shot Object Detection }
 There exists a source domain dataset \(D_{source}\) and a target domain dataset \(D_{target}\), which have classes \(C_{source}\) and \(C_{target}\) respectively, and we have \(C_{source}\cap C_{target}=\phi\). Typically, the model is initialized from a model pre-trained on \(D_{source}=\{x^s_{i}, y^s_{i}\}_{i = 1}^{N_S}\) which is used for acquiring source-domain knowledge where \(y^s_{i}\in C_{source}\). Then, the object detection model obtained after training on the source-domain dataset is transferred to the target-domain dataset \(D_{target}=\{x^t_{i}, y^t_{i}\}_{i = 1}^{N_T}\)where \(y^t_{i}\in C_{target}\). For each target-domain class, there are only 1, 5, or 10 training samples, which makes the fine-tuning training in the target domain challenging. For fairness, in the target-domain fine-tuning stage (our main focus), we adopt the setting of ~\cite{fu2024cross}, with the support set \(S = \{x_S, y_S\}\) and the query set fixed. The number of classes in each target-domain dataset is fixed, and each class has only 1, 5, or 10 samples. After fine-tuning on the support set (our main focus), the model is evaluated on the query set $Q=\{x_Q,y_Q\}$. 
\subsubsection{Generative Models}
The core objective of generative models is to learn the underlying probability distribution $p_\theta(x)$ of real data $x\sim p_{data}(x)$, and to be able to sample new data $x_{new}$ from it~\cite{luo2022understanding}. Meanwhile, it ensures that the generated samples are both of high quality (realistic) and can cover the diverse patterns of the data distribution.
In this work, firstly, we input the original image into a large language model (LLM) to obtain a textual description of the image, and then use existing generative models to generate images via a text-and-image-to-image process.

\vspace{-0.3cm}

\subsubsection{Finetuning}
In the fine-tuning phase, we specifically optimize the parameters of the detection head, classification head, and newly added modules for the target domain support set while keeping the backbone network frozen. Finally, the model performance is evaluated on the target domain test set. In this work, we use CD-ViTO~\cite{fu2024cross} for fine-tuning in experiments. Our method is general and also compatible with alternative fine-tuning schemes. 
\begin{figure}[t]
    \centering
    \includegraphics[width=0.6\linewidth]{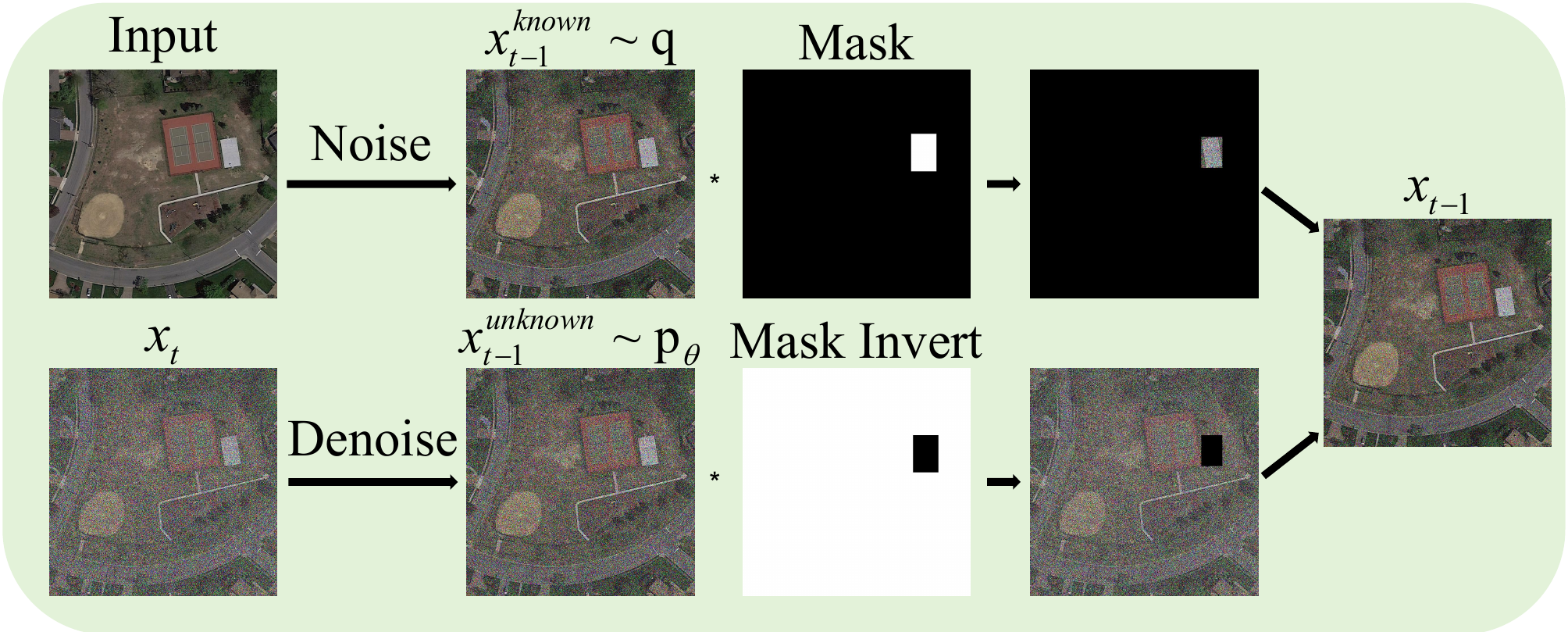}
    \caption{
    Inpainting at timestep $t$, where the intermediate result $x_{t-1} ^{known}$ from the forward process and the prediction $x_{t-1} ^{unknown}$ from the reverse process are jointly synthesized to generate the image $x_{t-1}$ in the backward process.
    }
    
    \label{fig: Inpaint}
    \vspace{-0.5cm}
    
\end{figure}

\vspace{-0.3cm}

\subsection{Inpainting in Diffusion Models}
\vspace{-0.2cm}

Inpainting in diffusion models fills in missing, occluded, or damaged areas in an image~\cite{lugmayr2022repaint}, which involves progressively reconstructing the missing parts while preserving known regions through the backward process of diffusion models (As shown in Fig.~\ref{fig: Inpaint}). Specifically, the model first corrupts the intact image by adding noise. During the reverse denoising process, it treats the undamaged areas as conditional constraints and guides the generation of missing regions through mechanisms like cross-attention. 
At each denoising step, the system enforces the retention of known pixels while predicting exclusively within the masked areas, ensuring semantic, structural, and textural consistency between the generated content and its surroundings.

\begin{figure*}[t]
    \centering
    \includegraphics[width=0.9\linewidth]{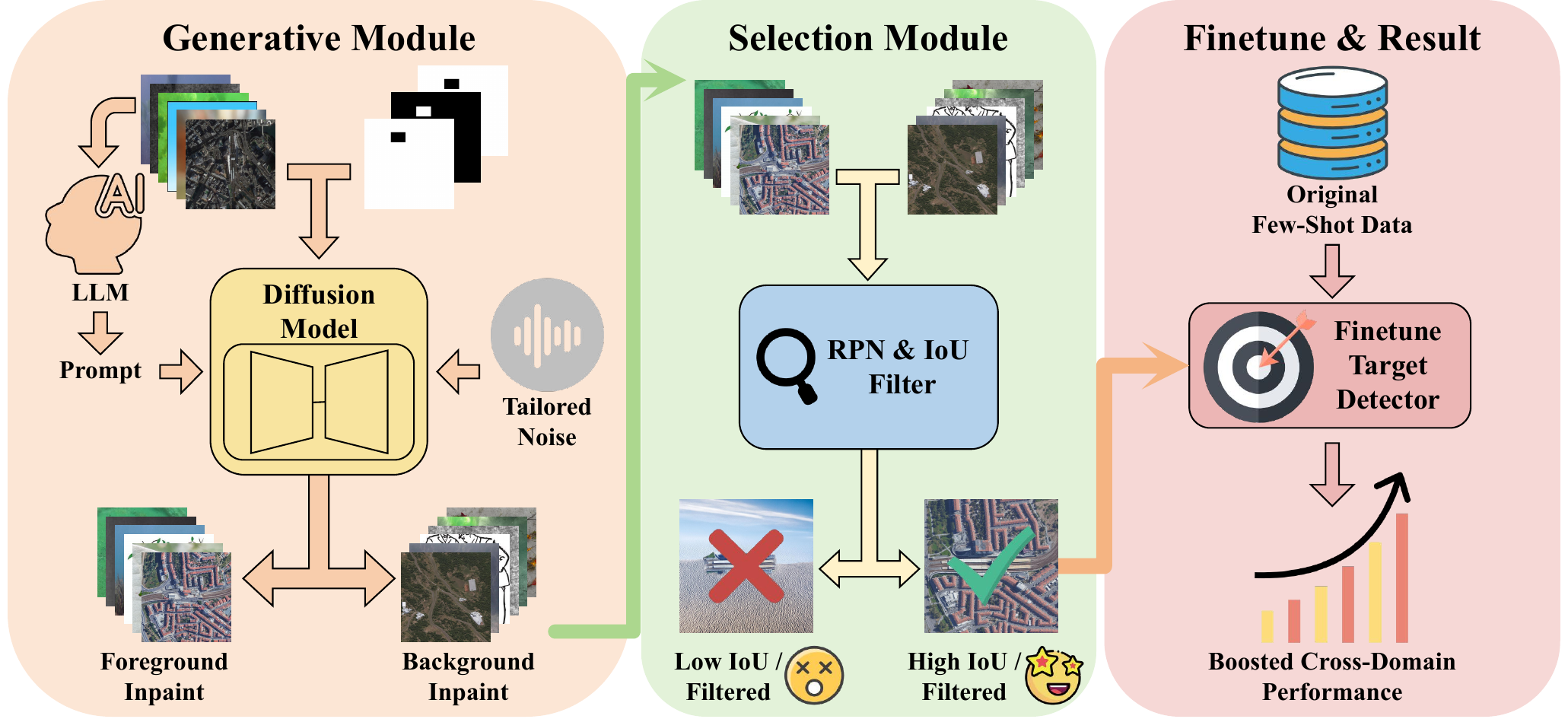}
    \caption{Our method operates solely on the target domain and consists of three stages. In the first stage, we feed images into an LLM to obtain prompts, extract bounding boxes from annotations, and compute their masks. In the second stage, the prompts, original images, and bounding box information are input into a diffusion model to perform inpainting on both foreground and background regions of the images. During the third stage, the generated images are fed into an encoder to calculate Intersection over Union (IoU) metrics. Through a selection module, we select higher-quality samples to be incorporated into the target detection model for target-domain finetuning.}
    \label{fig: Method}
    \vspace{-0.5cm}
    
\end{figure*}

\vspace{-0.3cm}

\subsection{Selective Inpainting with Tailored Noises}  
\vspace{-0.2cm}

SITN operates through three sequential stages: initially obtaining generative samples via the Generation Module, subsequently filtering the generated samples through the Selection Module to construct an expanded support set as shown in Fig.~\ref{fig: Method}, and ultimately applying it to downstream object detection tasks for both training and inference purposes. 

\vspace{-0.3cm}

\subsubsection{Generation Module}
As in Section 2, the style of target-domain images significantly differs from the source domain, with backgrounds being more transferable than foregrounds. 
Therefore, apart from semantic regions of images, we also mask the background regions with tailored noise and use diffusion models to inpaint these regions, which is used as the main augmentation.
The specific approach is as follows:

1. Semantic Description Generation: An LLM extracts a global description of original support-set images $x_{sup}$ as the input $prompt$ for the diffusion model.
\vspace{-0.2cm}
\begin{equation}
    prompt = M_{LLM}(x_{sup})
\end{equation}
\vspace{-0.4cm}

2. Target Region Localization: Based on ground truth annotations from the dataset, the object detection bounding box (BBox) is extracted as the mask for inpainting.
\vspace{-0.2cm}
\begin{equation}
mask_{sup} = 
\begin{cases}
1, & \text{if the pixel is in } y_{sup}\\
0, & \text{if the pixel is not in } y_{sup}
\end{cases}
\end{equation}

3. Diffusion Model Synthesis: The prompt, original image, and mask are sent into a frozen diffusion model to synthesize new images that are visually and semantically similar to the original. For noise control, we set the maximum step $T_{max}=1000$ and \textbf{adjust the actual noise step} $T=\epsilon T_{max}$ using a noise strength scalar $\epsilon < 1$, preventing $x_T$ from degenerating entirely into Gaussian noise.
\vspace{-0.5cm}

\begin{equation}
x_{sup}^{low}=AddNoise(x_{sup},\epsilon_{low})
\end{equation}

\begin{figure}[t]
    \centering
    \includegraphics[width=0.6\linewidth]{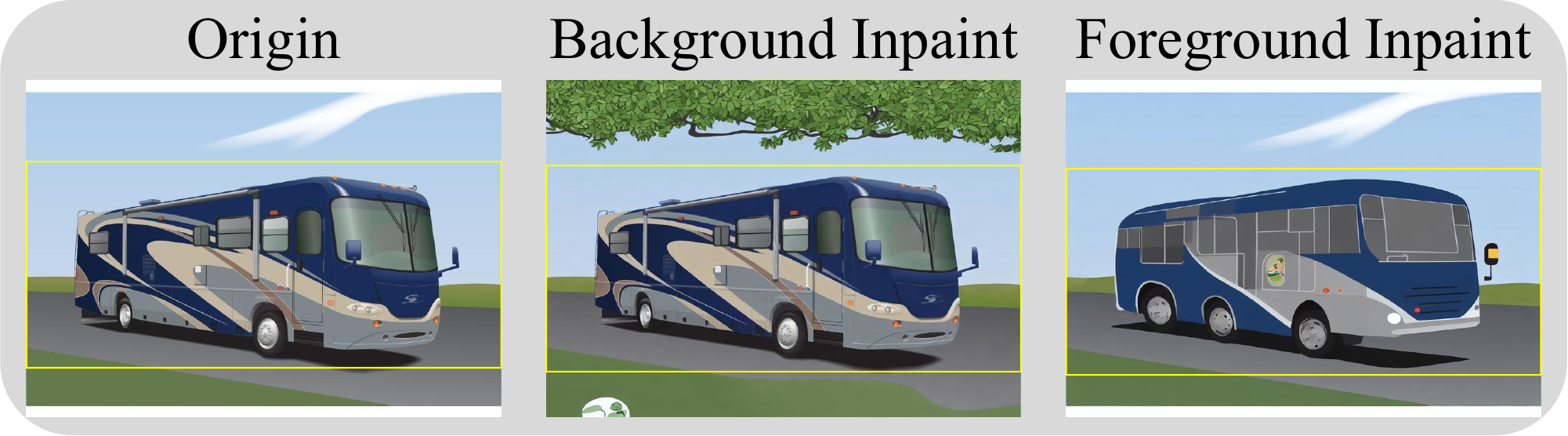}
    \vspace{-0.3cm}
    \caption{Images generated by foreground and background inpainting, with the bounding box (yellow).}

    \label{fig: without requiring re-annotation}
    \vspace{-0.5cm}
    
\end{figure}

During this process, the diffusion model leverages knowledge pre-trained on the source domain to infer semantic information obscured by noise, effectively restoring key content. Thus, the denoised foreground \(x_{sup}^{fore}\)  and background \(x_{sup}^{back}\)  remain category-consistent with the original image \(x_{sup}\). Moreover, since the mask is determined by the BBox, the inpainted images retain the original object positions without requiring re-annotation(As shown in Fig.~\ref{fig: without requiring re-annotation}).
\begin{equation}
x_{sup}^{back} = M_{diff}(x_{sup}^{low}, \text{prompt}, mask_{sup})
\end{equation}
\vspace{-0.5cm}
\begin{equation}
x_{sup}^{fore} = M_{diff}(x_{sup}^{low}, \text{prompt} + \text{category}, (1 - mask_{sup}))
\end{equation}
Finally, the augmented image-label pairs \(\{x_{sup}^{back}, y_{sup}\}\) and \(\{x_{sup}^{fore}, y_{sup}\}\) are combined with the original support set $S$ to form two extended support sets.

\subsubsection{Selection Module}

However, we also observe that not all generated samples are effective for augmentation, which is possibly caused by the ineffective LLM-generated descriptions or non-jointly trained LLM and diffusion models. Moreover, we also observe that some foreground-inpainted images are also helpful for augmentation, although the ratio of useful samples is much smaller than the proposed background-inpainted ones.
Therefore, to further improve the quality of generated samples for CDFSOD, we propose a selection module to further filter out ineffective samples in the extended support sets. Finally, the two extended sets are simultaneously sent into the object detection model.

Specifically, for each candidate image, we compute the box IoU (Intersection over Union) values between the enhanced images \(x_{sup}^{back}\)  and \(x_{sup}^{fore}\), and then select the positive samples based on IoU values. These samples are then used for subsequent tasks in object detection, including feature extraction, instance reweighting, domain prompting, fine-tuning, and inference.

We employ the Region Proposal Network (RPN) pre-trained on the CD-ViTO model (Baseline) to generate candidate bounding boxes ($A$) for the extended support set. We calculate the average IoU between predicted boxes and ground truth ones ($Y$), and only synthesized images with high IoUs are selected.
\begin{equation}
    x_{select}^{k} = M_{RPN}^{k}\Sigma(x_{sup}^{back},x_{sup}^{fore})
\end{equation}

We employ the top-k selection algorithm to filter the generated images, with the value of k tailored to each target domain dataset. The specific choices of k are detailed in the supplementary material.

\begin{table}[t]
  \centering
  
  \caption{The main results including the state-of-the-art works and works with ours method under 1-shot, 5-shot, 10-shot settings (mAP). $\dagger$ means results are produced by us. Avg. means average result. Values in the table are highlighted as follows: bold for the best and underlined for the second best, respectively.}
  \vspace{-0.3cm}
    \resizebox{0.8\textwidth}{!}{
    \begin{tabular}{clcccccccc}
    \toprule
    Setting & Method & Backbone & ArTaxOr & Clipart1k & DIOR  & DeepFish & NEU-DET & UODD  & Avg. \\
    \midrule
          & CDMM-FSOD & DETR-R101 & 15.1  & -     & 14.3  & -     & 6.9 & -     & / \\
          & \cellcolor{cyan!10}CDMM-FSOD + \textbf{Ours}$\dagger$ & \cellcolor{cyan!10}DETR-R101 & \cellcolor{cyan!10}16.3  & \cellcolor{cyan!10}-     & \cellcolor{cyan!10}15.1  & \cellcolor{cyan!10}-     & \cellcolor{cyan!10}7.4 & \cellcolor{cyan!10}-     & \cellcolor{cyan!10}/ \\
          & ViTDeT-FT & ViT-B/14 & 5.9   & 6.1   & 12.9  & 0.9   & 2.4   & 4.0   & 5.4  \\
          & \cellcolor{cyan!10}ViTDeT-FT + \textbf{Ours}$\dagger$ & \cellcolor{cyan!10}ViT-B/14 & \cellcolor{cyan!10}5.4  & \cellcolor{cyan!10}7.2   & \cellcolor{cyan!10}9.6  & \cellcolor{cyan!10}5.1 & \cellcolor{cyan!10}6.3  & \cellcolor{cyan!10}5.6  & \cellcolor{cyan!10}6.5 \\
    1-shot& DE-ViT-FT & ViT-L/14 & 10.5  & 13.0  & 14.7  & 19.3  & 0.6   & 2.4   & 10.1  \\
          & \cellcolor{cyan!10}DE-ViT-FT + \textbf{Ours}$\dagger$ & \cellcolor{cyan!10}ViT-L/14 & \cellcolor{cyan!10}12.6  & \cellcolor{cyan!10}14.1 & \cellcolor{cyan!10}17.1  & \cellcolor{cyan!10}20.2  & \cellcolor{cyan!10}2.5  & \cellcolor{cyan!10}3.2   & \cellcolor{cyan!10}11.6  \\
          & CD-ViTO & ViT-L/14 & 21.0  & 17.7  & 17.8  & 20.3  & 3.6   & 3.1   & 13.9  \\
          & \cellcolor{cyan!10}CD-ViTO + \textbf{Ours}$\dagger$  & \cellcolor{cyan!10}ViT-L/14 & \cellcolor{cyan!10}26.7 & \cellcolor{cyan!10}22.3 & \cellcolor{cyan!10}\textbf{19.8} & \cellcolor{cyan!10}22.1 & \cellcolor{cyan!10}5.4   & \cellcolor{cyan!10}5.6   & \cellcolor{cyan!10}17.0 \\
          & GLIP$\dagger$  & Swin-B & 12.6  & 52.1 & 4.8  & 40.1  & 1.6   & 4.5   & 19.3  \\
          & \cellcolor{cyan!10}GLIP + \textbf{Ours}$\dagger$  & \cellcolor{cyan!10}Swin-B & \cellcolor{cyan!10}38.4 & \cellcolor{cyan!10}54.7 & \cellcolor{cyan!10}16.8 & \cellcolor{cyan!10}\underline{41.3} & \cellcolor{cyan!10}3.2   &  \cellcolor{cyan!10}4.6  & \cellcolor{cyan!10}26.3 \\
        
          & ETS$\dagger$   & Swin-B & 13.7  & 51.0 & 7.4  & 35.0  & 7.0   & 11.0   & 21.9  \\
          & \cellcolor{cyan!10}ETS + \textbf{Ours}$\dagger$  & \cellcolor{cyan!10}Swin-B & \cellcolor{cyan!10}51.0 & \cellcolor{cyan!10}52.7 & \cellcolor{cyan!10}15.6 & \cellcolor{cyan!10}\textbf{43.2} & \cellcolor{cyan!10}\textbf{13.9}  &  \cellcolor{cyan!10}19.3  & \cellcolor{cyan!10}32.1 \\
          & GroundDINO$\dagger$   & Swin-B & 20.0  & \underline{57.6}  & 8.0  & 34.3  & 7.2   & 17.1 & 24.0  \\
           & DomainRAG   & Swin-B & \textbf{57.2}  & 56.1  & \underline{18.0}  & 38.0  & 12.1   & \underline{20.2} & \underline{33.6}  \\
        & \cellcolor{cyan!10}GroundDINO + \textbf{Ours}$\dagger$  & \cellcolor{cyan!10}Swin-B & \cellcolor{cyan!10}\underline{52.3} & \cellcolor{cyan!10}\textbf{59.0} & \cellcolor{cyan!10}16.5 & \cellcolor{cyan!10}\underline{41.3} & \cellcolor{cyan!10}\underline{13.6}  &  \cellcolor{cyan!10}\textbf{22.8}  & \cellcolor{cyan!10}\textbf{34.3} \\
    \midrule
          & CDMM-FSOD & DETR-R101 & 48.7  & -     & 26.9  & -     & 12.5  & -     & / \\
          & \cellcolor{cyan!10}CDMM-FSOD + \textbf{Ours}$\dagger$ & \cellcolor{cyan!10}DETR-R101 & \cellcolor{cyan!10}55.0  & \cellcolor{cyan!10}-     & \cellcolor{cyan!10}29.1  & \cellcolor{cyan!10}-     & \cellcolor{cyan!10}17.5 & \cellcolor{cyan!10}-     & \cellcolor{cyan!10}/ \\
          & ViTDeT-FT & ViT-B/14 & 20.9  & 23.3  & 23.3  & 9.0   & 13.5  & 11.1  & 16.9  \\
           & \cellcolor{cyan!10}ViTDeT-FT$\dagger$ + \textbf{Ours} & \cellcolor{cyan!10}ViT-B/14 & \cellcolor{cyan!10}18.0  & \cellcolor{cyan!10}25.1  & \cellcolor{cyan!10}21.1  & \cellcolor{cyan!10}18.8 & \cellcolor{cyan!10}17.1   & \cellcolor{cyan!10}13.5  & \cellcolor{cyan!10}19.2 \\
    5-shot& DE-ViT-FT & ViT-L/14 & 38.0  & 38.1  & 23.4  & 21.2  & 7.8   & 5.0   & 22.3  \\
         & \cellcolor{cyan!10}DE-ViT-FT + \textbf{Ours}$\dagger$ & \cellcolor{cyan!10}ViT-L/14 & \cellcolor{cyan!10}39.5  & \cellcolor{cyan!10}41.2 & \cellcolor{cyan!10}22.9  & \cellcolor{cyan!10}22.9  & \cellcolor{cyan!10}8.1  & \cellcolor{cyan!10}6.1   & \cellcolor{cyan!10}23.5  \\
          & CD-ViTO & ViT-L/14 & 47.9  & 41.1  & 26.9  & 22.3  & 11.4  & 6.8   & 26.1  \\
          & \cellcolor{cyan!10}CD-ViTO + \textbf{Ours}$\dagger$  & \cellcolor{cyan!10}ViT-L/14 & \cellcolor{cyan!10}\underline{52.3} & \cellcolor{cyan!10}42.8 & \cellcolor{cyan!10}27.2 & \cellcolor{cyan!10}22.6  & \cellcolor{cyan!10}13.4  & \cellcolor{cyan!10}7.6   & \cellcolor{cyan!10}27.7 \\
          & GLIP$\dagger$  & Swin-B & 15.0  & 57.5  & 11.8  & 41.5  & 9.4   & 4.4   & 23.3  \\
          & \cellcolor{cyan!10}GLIP + \textbf{Ours}$\dagger$  & \cellcolor{cyan!10}Swin-B & \cellcolor{cyan!10}16.9 & \cellcolor{cyan!10}58.9 & \cellcolor{cyan!10}13.1 & \cellcolor{cyan!10}41.4 &  \cellcolor{cyan!10}16.0  & \cellcolor{cyan!10}4.4  &  \cellcolor{cyan!10}25.1 \\

          & ETS$\dagger$   & Swin-B & 33.0  & 58.4  & 12.8  & 42.1  & 9.2   & 19.3  & 29.1  \\
          & \cellcolor{cyan!10}ETS + \textbf{Ours}$\dagger$  & \cellcolor{cyan!10}Swin-B & \cellcolor{cyan!10}63.2 & \cellcolor{cyan!10}\textbf{61.1} & \cellcolor{cyan!10}22.2 & \cellcolor{cyan!10}\textbf{51.1} &  \cellcolor{cyan!10}15.1  &  \cellcolor{cyan!10}\underline{27.4}  & \cellcolor{cyan!10}40.0 \\
           & GroundDINO$\dagger$   & Swin-B & 29.0 & \underline{60.1}  & 16.0  & 42.3  & 12.5   & 25.5 & 30.9\\
           & DomainRAG   & Swin-B & \textbf{70.0}  & 59.8  & \textbf{31.5}  & 43.8  & \textbf{24.2}   & 26.8 & \underline{42.7}  \\
        & \cellcolor{cyan!10}GroundDINO + \textbf{Ours}$\dagger$  & \cellcolor{cyan!10}Swin-B & \cellcolor{cyan!10}\underline{65.6} & \cellcolor{cyan!10}\textbf{61.1} & \cellcolor{cyan!10}\underline{29.8} & \cellcolor{cyan!10}\underline{50.5} & \cellcolor{cyan!10}\underline{23.4}  &  \cellcolor{cyan!10}\textbf{31.7} & \cellcolor{cyan!10}\textbf{43.7} \\
    \midrule
          & CDMM-FSOD & DETR-R101 & 61.4  & -     & 31.4  & -     & 17.5  & -     & / \\
          & \cellcolor{cyan!10}CDMM-FSOD + \textbf{Ours}$\dagger$ & \cellcolor{cyan!10}DETR-R101 & \cellcolor{cyan!10}63.1  & \cellcolor{cyan!10}-     & \cellcolor{cyan!10}33.2  & \cellcolor{cyan!10}-     & \cellcolor{cyan!10}18.6 & \cellcolor{cyan!10}-     & \cellcolor{cyan!10}/ \\
          & ViTDeT-FT & ViT-B/14 & 23.4  & 25.6  & 29.4  & 6.5   & 15.8  & 15.6 & 19.4  \\
          & \cellcolor{cyan!10}ViTDeT-FT + \textbf{Ours}$\dagger$ & \cellcolor{cyan!10}ViT-B/14 & \cellcolor{cyan!10}20.7  & \cellcolor{cyan!10}27.2   & \cellcolor{cyan!10}27.2  & \cellcolor{cyan!10}21.2 & \cellcolor{cyan!10}20.1   & \cellcolor{cyan!10}20.1  & \cellcolor{cyan!10}22.7 \\
    10-shot & DE-ViT-FT & ViT-L/14 & 49.2  & 40.8  & 25.6  & 21.3  & 8.8   & 5.4   & 25.2  \\
          & \cellcolor{cyan!10}DE-ViT-FT + \textbf{Ours}$\dagger$ & \cellcolor{cyan!10}ViT-L/14 & \cellcolor{cyan!10}\textbf{51.1}  & \cellcolor{cyan!10}42.2 & \cellcolor{cyan!10}26.3  & \cellcolor{cyan!10}22.5  & \cellcolor{cyan!10}9.6  & \cellcolor{cyan!10}7.2   & \cellcolor{cyan!10}26.5  \\
          & CD-ViTO & ViT-L/14 & 60.5  & 44.3  & 30.8  & 22.3  & 12.8  & 7.0   & 29.6  \\
          & \cellcolor{cyan!10}CD-ViTO + \textbf{Ours}$\dagger$  & \cellcolor{cyan!10}ViT-L/14 & \cellcolor{cyan!10}61.4 & \cellcolor{cyan!10}46.5 & \cellcolor{cyan!10}31.9 & \cellcolor{cyan!10}23.6  & \cellcolor{cyan!10}15.2  & \cellcolor{cyan!10}9.3   & \cellcolor{cyan!10}31.3 \\
          & GLIP$\dagger$  & Swin-B & 26.3  & 55.3  & 14.8  & 36.4  & 9.3   & 15.9   & 26.3  \\
          & \cellcolor{cyan!10}GLIP + \textbf{Ours}$\dagger$  & \cellcolor{cyan!10}Swin-B & \cellcolor{cyan!10}49.9 & \cellcolor{cyan!10}60.5 & \cellcolor{cyan!10}24.2 & \cellcolor{cyan!10}42.0 &  \cellcolor{cyan!10}8.7  & \cellcolor{cyan!10}6.5  & \cellcolor{cyan!10}32.0 \\
          & ETS$\dagger$   & Swin-B & 63.1  & 61.2  & 30.8  & 40.1  & 18.9   & 21.5   &  39.3\\
          & \cellcolor{cyan!10}ETS + \textbf{Ours}$\dagger$  & \cellcolor{cyan!10}Swin-B & \cellcolor{cyan!10}\textbf{74.1} & \cellcolor{cyan!10}\underline{62.7}& \cellcolor{cyan!10}\underline{37.8} &\cellcolor{cyan!10}\underline{47.9} &  \cellcolor{cyan!10}22.5 &  \cellcolor{cyan!10}25.8  & \cellcolor{cyan!10}45.1 \\
           & GroundDINO$\dagger$   & Swin-B & 62.8 & 62.3  & 24.1  & 39.3  & 16.3   & 25.1 & 38.1\\
        & DomainRAG   & Swin-B & \underline{73.4}  & 61.1  & \textbf{39.0} & 41.3  & \textbf{26.3}   & \underline{31.2} & \underline{45.4}  \\   
        & \cellcolor{cyan!10}GroundDINO + \textbf{Ours}$\dagger$  & \cellcolor{cyan!10}Swin-B & \cellcolor{cyan!10}69.4 & \cellcolor{cyan!10}\textbf{65.2}& \cellcolor{cyan!10}32.6 & \cellcolor{cyan!10}\textbf{53.5} & \cellcolor{cyan!10}\underline{24.6}  &  \cellcolor{cyan!10}\textbf{34.8} & \cellcolor{cyan!10}\textbf{46.9} \\
    \bottomrule

    \end{tabular}%

    }
        \vspace{-0.4cm}
  \label{tab: Sota}%
  
\end{table}%

\section{Experiments}

\vspace{-0.3cm}

\subsection{Dataset and evaluation setup}
\vspace{-0.2cm}
We use the benchmark of~\cite{fu2024cross} to conduct experimental evaluations for Cross-Domain Few-Shot Object Detection. The model is trained on the COCO dataset and evaluated on six datasets, namely ArTaxOr, DIOR, UODD, Clipart1k, DeepFish, and NEU-DET.

\begin{figure}[t]
    \centering
    
    \includegraphics[width=1.0\linewidth]{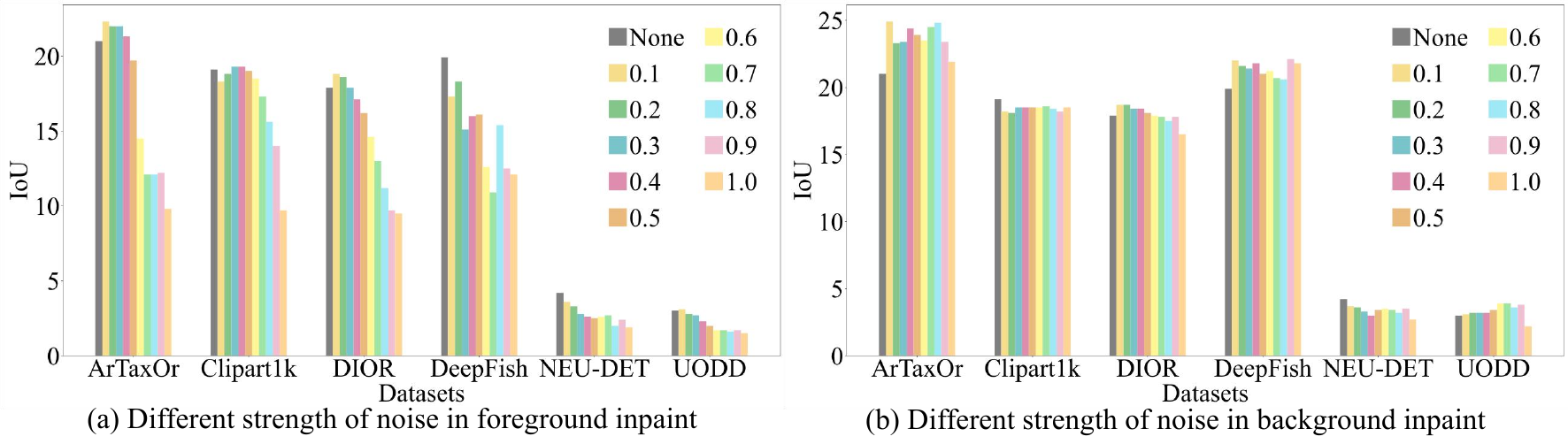}
    \caption{Ablation study of tailored noise. The legends in different colors indicate different noise strengths.}
    \label{fig: Ablation study of Generation Module}
    
\end{figure}

\vspace{-0.6cm}
\subsection{Implementation Details}
   \vspace{-0.2cm}

We evaluate model performance using mean Average Precision (mAP) under 1-shot, 5-shot, and 10-shot settings. For all generated datasets, the experimental pipeline adheres to the three-phase ``pretrain, fine-tune, test" protocol. The implementation first initializes with a DE-ViT backbone pre-trained on the COCO dataset, followed by module-specific optimization leveraging the expanded support set. Comprehensive finetuning configurations—hyperparameter settings, optimizer selection, learning rate schedules, training loss, and training epochs—are in the appendix.

\subsection{Comparison with state-of-the-art works}
\vspace{-0.2cm}

Tab.~\ref{tab: Sota} presents a comprehensive comparison with state-of-the-art methods, evaluating three backbone architectures: Swin-B, Vision Transformer-L (ViT-L), Vision Transformer-B (ViT-B), and DETR-R101, under 1-shot, 5-shot, and 10-shot configurations. The benchmark approaches include CDMM-FSOD~\cite{shangguan2025cross}, ViTDeT-FT~\cite{li2022exploring}, DE-ViT-FT~\cite{zhang2023detect}, CD-ViTO (baseline)~\cite{fu2024cross},  
GLIP~\cite{li2022grounded}, ETS~\cite{pan2025enhance}, GroundDINO~\cite{liu2023grounding}, DomainRAG~\cite{li2025domain}, series. We consistently achieve state-of-the-art average performance in all settings.

\subsection{Ablation Study}

\vspace{-0.2cm}
\subsubsection{Generative Module}

(1) Foreground inpainting: As analyzed in Section 3, within specific noise strength ($\epsilon$) parameter ranges, the Generation Module can effectively synthesize qualified foreground samples. Fig.~\ref{fig: Ablation study of Generation Module} (a) demonstrates the existence of a definable parameter interval that can produce foreground inpainting samples suitable for the CDFSOD task.

(2) Background inpainting: Fig.~\ref{fig: Ablation study of Generation Module} (b) shows that even without relying on the Section Module's filtering mechanism, a significant proportion of samples independently generated by the background inpainting could still enhance the performance in the CDFSOD task. This phenomenon indicates that the background inpainting itself possesses intrinsic effectiveness, with its performance improvement not entirely dependent on subsequent selection strategies.

(3) Other diffusion models: Tab.~\ref{tab: Ablation study of other diffusion models} indicates that different versions of the Stable-Diffusion (including SDv2 and SDv3) achieve comparable performance in terms of visual fidelity of the generated images. However, there exists a noticeable quality gap between these versions and other existing diffusion-based methods (including Diffmix~\cite{islam2024diffusemix}, IMBABM~\cite{he2022synthetic}, DA-fusion~\cite{trabucco2023effective}, SDEdit~\cite{meng2021sdedit}, ControlNET~\cite{zhang2023adding}, and IP-Adapter~\cite{ye2023ip}).

\begin{table*}[t]
  \centering
   \caption{Comparison with other diffusion models. $\dagger$ means results are produced by us.}
   \resizebox{0.6\textwidth}{!}{
    \begin{tabular}{cccccccc}
    \hline
    Method & ArTa. & Clip. & DIOR  & Deep. & NEU.  & UODD  & Avg. \\
    \hline
    Diffmix† & 11.9  & 6.7   & 7.3   & 0.1   & 2.4   & 0.2   & 4.8 \\
    IMBABM† & 5.9   & 3.5   & 3.9   & 5.2   & 0.9   & 0.9   & 3.4 \\
    Da-fusion† & 18.7  & 18.5  & 14.7  & 15.8  & 2.4   & 1.9   & 12.0  \\
    SDEdit† & 35.5  & 41.5  & 13.1  & 35.7  & 9.8   & 13.7  & 13.2 \\
    IP-Adapter† & 2.5   & 15.4  & 7.1   & 1.1   & 3.5   & 1.0     & 5.1 \\
    ControlNet† & 3.1   & 23.3  & 7.5   & 24.5  & 7.2   & 3.0     & 11.4 \\
    Ours† & \textbf{52.3} & \textbf{59.0} & \textbf{16.5} & \textbf{41.3} & \textbf{13.6} & \textbf{22.8} & \textbf{34.3} \\
    \hline
    \end{tabular}
    }
    \vspace{-0.2cm}    
  \label{tab: Ablation study of other diffusion models}
\end{table*}

\subsubsection{Selection Module}

The effectiveness of Selection: As shown in Tab.~\ref{tab: Ablation study of Selection Module}, the performance improvement achieved solely by the baseline model without the Selection Module is relatively limited. This is possibly caused by ineffective image descriptions and non-jointly trained LLM and diffusion models, resulting in significant quality variations in the images directly generated by the Generation Module—some samples are difficult for object detection models to identify accurately. Therefore, our proposed Selection Module effectively filters high-quality generated images. These carefully selected samples, combined with the original training data for object detection model finetuning, lead to a substantial overall performance enhancement. 
\begin{table}[t]
  \centering
  \caption{Ablation study of Selection Module (mAP). Fore. means that only the foreground is synthesized. Back. means that only the background is synthesized. Sele. means Selection Module.}
  \vspace{-0.3cm}
  \resizebox{0.7\textwidth}{!}{
    \begin{tabular}{lccccc}
    \toprule
    Dataset & Baseline & Fore. & Back. & Fore. + Back. & Fore. + Back. + Sele. \\
    \midrule
    ArTa. & 21.0  & 22.3  & 24.9 & 22.5 & \textbf{26.7} \\
    Clip. & 17.7  & 19.3  & 18.6 & 19.3 & \textbf{22.3} \\
    DIOR  & 17.8  & 18.8  & 18.7 & 16.6 & \textbf{19.8} \\
    Deep.  & 20.3  & 18.3  & 22.1 & 13.5 & \textbf{22.1} \\
    NEU. & 3.6   & 3.6   & 3.7  & 3.2 & \textbf{5.4} \\
    UODD  & 3.1   & 3.1   & 3.9  & 2.1 & \textbf{5.6} \\
    \bottomrule
    \end{tabular}
    }
    \vspace{-0.5cm}
    
\label{tab: Ablation study of Selection Module}
\end{table}



    

\vspace{-0.3cm}

\subsubsection{Migrating to CDFSS tasks }

As shown in Tab.~\ref{tab: on CDFSS}, our method has been successfully extended to the cross-domain few-shot segmentation (CDFSS) task, demonstrating competitive performance and validating its generalizability. The benchmark approaches include PerSAM~\cite{zhang2023personalize}, RePRI~\cite{boudiaf2021few}, HSNet~\cite{min2021hypercorrelation}, PATNet~\cite{lei2022cross}, SSP~\cite{fan2022self}, ABCDFSS~\cite{herzog2024adapt}, DRA~\cite{su2024domain}, APSeg~\cite{he2024apseg}, FPTrans~\cite{zhang2022feature}, LoEC~\cite{liu2025devil} series.

\begin{table}[h]

  \centering
  \vspace{-0.3cm}
  
  \caption{Migrating SITN to CDFSS tasks}
  \vspace{-0.3cm}
  \resizebox{1.0\textwidth}{!}{
    \begin{tabular}{ccccccccccccc}
    \hline
    Method & PerSAM & RePRI & HSNet & PATNet & SSP   & APM   & ABCDFSS & DRA   & APSeg & FPTrans & LoEC  & FPTrans+Ours \\
    \hline
    FSS-1000 & 60.9  & 71    & 77.5  & 78.6  & 78.9  & 79.3  & 74.6  & 79.1  & 79.7  & 80.7  & 81.1  & \textbf{84.5} \\
    Deepgloble & 36    & 25    & 29.7  & 37.9  & 40    & 40.9  & 42.6  & 41.3  & 35.9  & 38.4  & 42.1  & \textbf{43.8} \\
    ISIC  & 23.3  & 23.3  & 31.2  & 41.2  & 35.5  & 41.7  & 45.7  & 40.8  & 45.4  & 48.6  & 52.9  & \textbf{53.4} \\
    Chest X-ray & 30    & 65.1  & 51.9  & 66.6  & 74.4  & 78.3  & 79.8  & 82.4  & 84.1  & 80.9  & 83.9  & \textbf{84.9} \\
    Average & 37.6  & 46.1  & 47.6  & 56.1  & 57.2  & 60    & 60.7  & 60.9  & 61.3  & 62.2  & 65    & \textbf{66.7} \\
    \hline
    \end{tabular}%
    }
  \vspace{-0.8cm}
    
  \label{tab: on CDFSS}%
  
\end{table}%

\subsubsection{Comparison with data augmentations}

We compare our method with other data augmentations under the 1-shot setting, including Mixup, flipping, and brightness enhancement (Bright+), Simple Copy-Paste~\cite{ghiasi2021simple} (SCP), Entropy Maximization~\cite{chan2021entropy} (EM), and Dense outlier detection~\cite{bevandic2021dense} (Dod).
Notably, Tab.~\ref{tab: Ablation study of existing data augmentation methods} shows that most augmentation methods cannot improve the performance against using only the original images (CD-ViTO), verifying the difficulty in augmentation under large domain gaps.
The results show that our method achieves significant performance gains compared to other methods. This means that SITN is different from traditional augmentation methods, effectively solving the scarce data problem in CDFSOD.

\begin{table}[t]
  \centering
  
  \caption{Comparison with existing data augmentation (mAP). Bright+ means Bright Enhance, SCP means Simple Copy-Paste, EM means Entropy Maximization, and Dod means Dense outlier detection.}
   \resizebox{0.7\textwidth}{!}{
    \begin{tabular}{ccccccccc}
    \hline
    Method & Mixup & Flip  & Bright+ & SCP   & EME   & Dod   & Baseline & Ours \\
    \hline
    ArTa. & 13.4  & 20.5  & 20.5  & 24.2  & 15.6  & 25.8  & 21.0  & \textbf{26.7} \\
    Clip. & 12.3  & 17.7  & 17.9  & 17.5  & 7.8   & 15.4  & 17.7  & \textbf{22.3} \\
    DIOR  & 13    & 17.3  & 18.66 & 16.7  & 7.7   & 16.8  & 17.8  & \textbf{19.8} \\
    Deep. & 10.9  & 20.3  & 17.5  & 15.5  & 11    & 21.8  & 20.3  & \textbf{22.1} \\
    NEU.  & 1.2   & 4.1   & 3.1   & 1.0   & 0.6   & 2.0   & 3.6   & \textbf{5.4} \\
    UODD  & 1.5   & 3.2   & 2.9   & 2.3   & 0.5   & 3.2   & 3.1   & \textbf{5.6} \\
    Avg.  & 8.7   & 13.9  & 13.4  & 12.9  & 7.2   & 14.8  & 13.9  & \textbf{17.0 } \\
    \hline
    \end{tabular}%
    }
  \label{tab: Ablation study of existing data augmentation methods}
\end{table}

\vspace{-0.3cm}

\subsubsection{Object Detection under Occlusion}
\begin{figure}[htbp]
    \centering
    \vspace{-0.3cm}
    \begin{minipage}{0.7\linewidth}
        \centering
        \includegraphics[width=\linewidth]{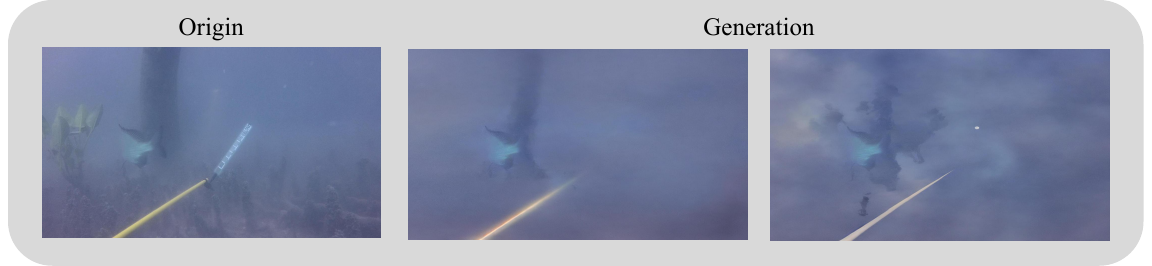}
    \caption{Application of the our method under occlusion dataset.}
        
        \label{fig: Occlusion}
    \end{minipage}
    \hfill
    \begin{minipage}{0.28\linewidth}
        \centering
        \resizebox{0.9\textwidth}{!}{
        \begin{tabular}{cc}
            \hline
            Dataset & Deep. \\
            \hline
            Baseline & 35.5 \\
            w/o Ours & 38.5 \\
            with Ours & \textbf{41.2} \\
            \hline
        \end{tabular}
        }
    \caption{Ablation study of occlusion scenes.}
        
        \label{tab: Occlusion}
    \end{minipage}
    
    \vspace{-0.3cm}
\end{figure}

\vspace{-0.3cm}
As shown in Fig.\ref{fig: Occlusion} and Fig.~\ref{tab: Occlusion}, we tested our method on a constructed 1-shot occlusion dataset and achieved promising results. Occlusion remains a general challenge in object detection, and our method offers a generic solution by supplementing data. Moreover, by adding controlled noise during generation, our approach preserves semantic information while improving diversity, making it effective for handling location variation, occlusion, and labeling errors.

\vspace{-0.5cm}


    
    

  
    

    

\subsubsection{Comparision with other open-source diffusion model}
As shown in Tab.~\ref{tab: open-source diffusion}, we use Nucleus-Image~\cite{nucleusimage2026} for generation. Directly applying Nucleus-Image yields poor performance. However, Nucleus-Image with our method performs comparably to SDv1.5 with our method. This shows our method remains effective on stronger diffusion models, confirming its long-term relevance.
    \vspace{-0.3cm}

\begin{table}[htbp]
  \centering
   \caption{Comparison with other open-source diffusion models. $\dagger$ means results are produced by us.}
  
  \resizebox{0.7\textwidth}{!}{
  
    \begin{tabular}{cccccccc}
    \hline
    Model & ArT.  & Clip. & DIOR  & Deep. & NEU.  & UODD  & Avg. \\
    \hline
    SDv1.5 & 13.9  & 11.3  & 8.6   & 6.2   & 2.5   & 1.7   & 7.4 \\
    Nucleus-Image$\dagger$ & 16.8     &  14.2    &  5.1    & 18.8     &  3.5    & 9.0     & 11.2 \\
    \hline
    Ours + SDv1.5 & 52.3  & 59    & 16.5  & \textbf{41.3 } & \textbf{13.6}  & \textbf{22.8} & \textbf{34.3} \\
    Ours + Nucleus-Image$\dagger$ & \textbf{55.4}  & \textbf{58.8}  & 16.3  & 36.9  & 8.4   & 16.8  & 32.1 \\
    \hline
    \end{tabular}%
    }
    \vspace{-0.4cm}
  \label{tab: open-source diffusion}%
\end{table}%

\vspace{-0.2cm}
\subsubsection{Computational Efficiency} 
\begin{figure*}[htbp]
\vspace{-0.2cm}
    \centering
    \begin{minipage}[c]{0.48\textwidth}
        \centering
        \includegraphics[width=1.0\linewidth]{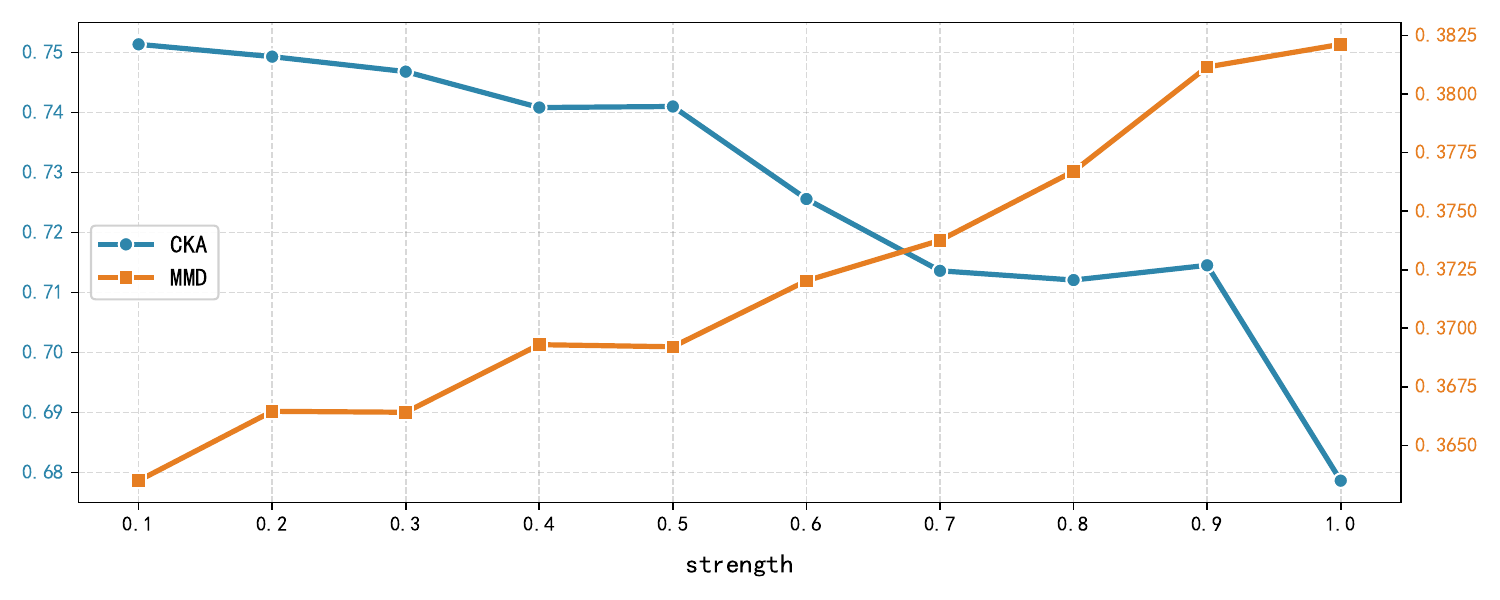}
        \caption{The variations of CKA and MMD.}
        \label{fig: CKA and MMD}
    \end{minipage}
    \hfill
    \begin{minipage}[c]{0.48\textwidth}
        \centering
        
        \resizebox{0.9\linewidth}{!}{%
        \begin{tabular}{ccc}
        \hline
             Method & Generation Time of 1 image \\
             \hline
             DDPM & 460s \\
             SDEdit & 300s-450s \\
             DomainRAG & 5s-8s \\
             Ours & 0.02s-1.5s \\
        \hline
        \end{tabular}%
        }
        \caption{The computational efficiency of the original diffusion model and our method.}
        \label{tab: computational efficiency}
       
    \end{minipage}
 \vspace{-0.8cm}
    
\end{figure*}

Due to the diffusion model only applying partial noise to the image, both the forward and backward processes avoid the complete diffusion steps; the image generation is constrained to within 50 steps by leveraging DDIM\cite{Song_Meng_Ermon_2020}. Moreover, our method is training-free. Fig.~\ref{tab: computational efficiency} shows the generation time of 1 image between the original diffusion model (DDPM)~\cite{Ho_Jain_Abbeel_2020} and our method.

\vspace{-0.2cm}
\subsubsection{CKA vs. MMD }
Moreover, the CKA similarity comparison is shown in Fig.~\ref{fig: Source vs Target}, where our method achieves much higher CKA scores than other diffusion-based methods, even approaching those of traditional augmentations. The comparison with diffusion-based methods in performance and generated samples is in Fig.~\ref{fig: placeholder}, where our method consistently achieves much higher performance.
We reproduce the CKA experiments by a widely-adopted metric, the MMD distance, where larger MMDs indicate larger domain distances and smaller CKAs. We can see the trend is perfectly symmetric in Fig.~\ref{fig: CKA and MMD}.

Additional experiments (e.g., comparisons between augmented and real samples, generation quantities) will be provided in the Appendix.

\section{Conclusion}
\vspace{-0.3cm}

We revisit a neglected method for the CDFSOD task: diffusion-based data augmentation, finding that domain gaps make it difficult for current diffusion-based methods to synthesize useful samples for supplementing the scarce training data. We then analyze the domain gap from both the visual and the semantic side, and propose our method, SITN, achieving state-of-the-art results.

\section*{Acknowledgments}
This work is supported by the National Natural Science Foundation of China under grants 62206102; the National Key Research and Development Program of China under grant 2024YFC3307900; the National Natural Science Foundation of China under grants 62436003, 62376103 and 62302184; Major Science and Technology Project of Hubei Province under grant 2025BAB011 and 2024BAA008; Hubei Science and Technology Talent Service Project under grant 2024DJC078; and Ant Group through CCF-Ant Research Fund. The computation is completed in the HPC Platform of Huazhong University of Science and Technology.


%
%
\bibliographystyle{splncs04}
\bibliography{main}
\end{document}


\title{	
Appendix for Free-Lunch Augmentation by Revisiting Diffusion-Based Data Generation for Cross-Domain Few-Shot Object Detection} 

\titlerunning{Abbreviated paper title}

\author{Zijian Zhuang\inst{\dagger} \and
Yixiong Zou\inst{\dagger}\thanks{Corresponding author, $\dagger~$Equal contribution} \and 
Yuhua Li \and
Ruixuan Li}

\authorrunning{Z. Zhuang, Y. Zou, et al.}

\institute{School of Computer Science and Technology, Huazhong University of Science and Technology, China \\
\email{\{zhuangzijian, yixiongz, idcliyuhua, rxli\}@hust.edu.cn}\\}

\maketitle

\section{Dataset Description}
\subsection{Source-domain Dataset}
\textbf{MS-COCO (Microsoft Common Objects in Context)}: It is a widely used benchmark dataset for object detection, instance segmentation, and image captioning. It contains over 330,000 images (with more than 200,000 labeled instances) across 80 object categories, spanning diverse everyday scenes and complex visual contexts.

\subsection{Target-domain Dataset}

\textbf{ArTaxOr}\cite{drange2019arthropod}: A specialized dataset focused on arthropod taxonomy, covering diverse classes such as insects, spiders, crustaceans, and millipedes, designed to support fine-grained biological classification tasks.

\begin{figure}[htbp]
    \centering
    \begin{minipage}[t]{0.48\textwidth}
        \centering
        \includegraphics[width=\linewidth]{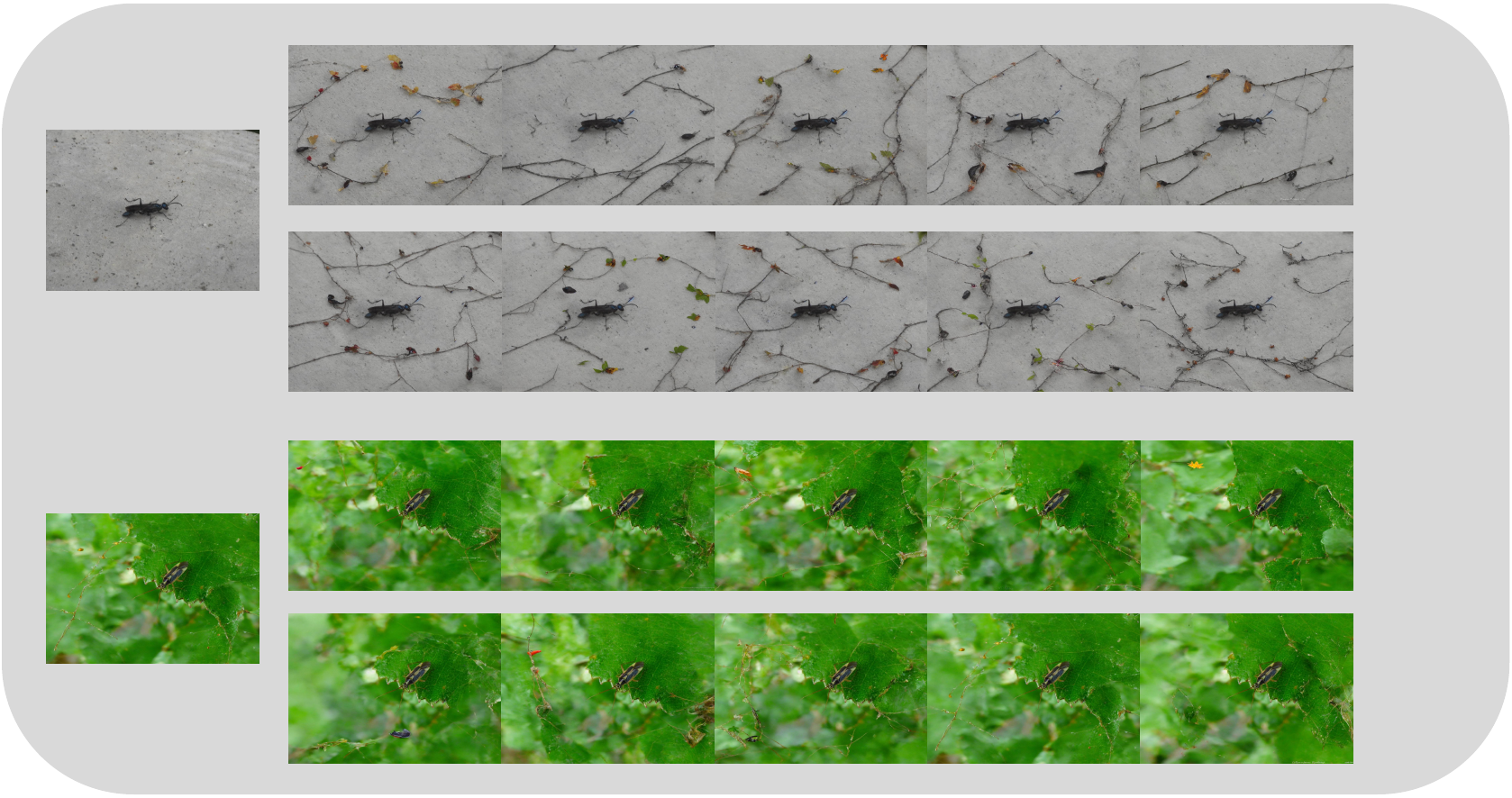}
        \caption{(left) The original image of the ArTaxOr. (right) The generated image of ArTaxOr.}
        \label{fig:artaxor}
    \end{minipage}
    \hfill
    \begin{minipage}[t]{0.48\textwidth}
        \centering
        \includegraphics[width=\linewidth]{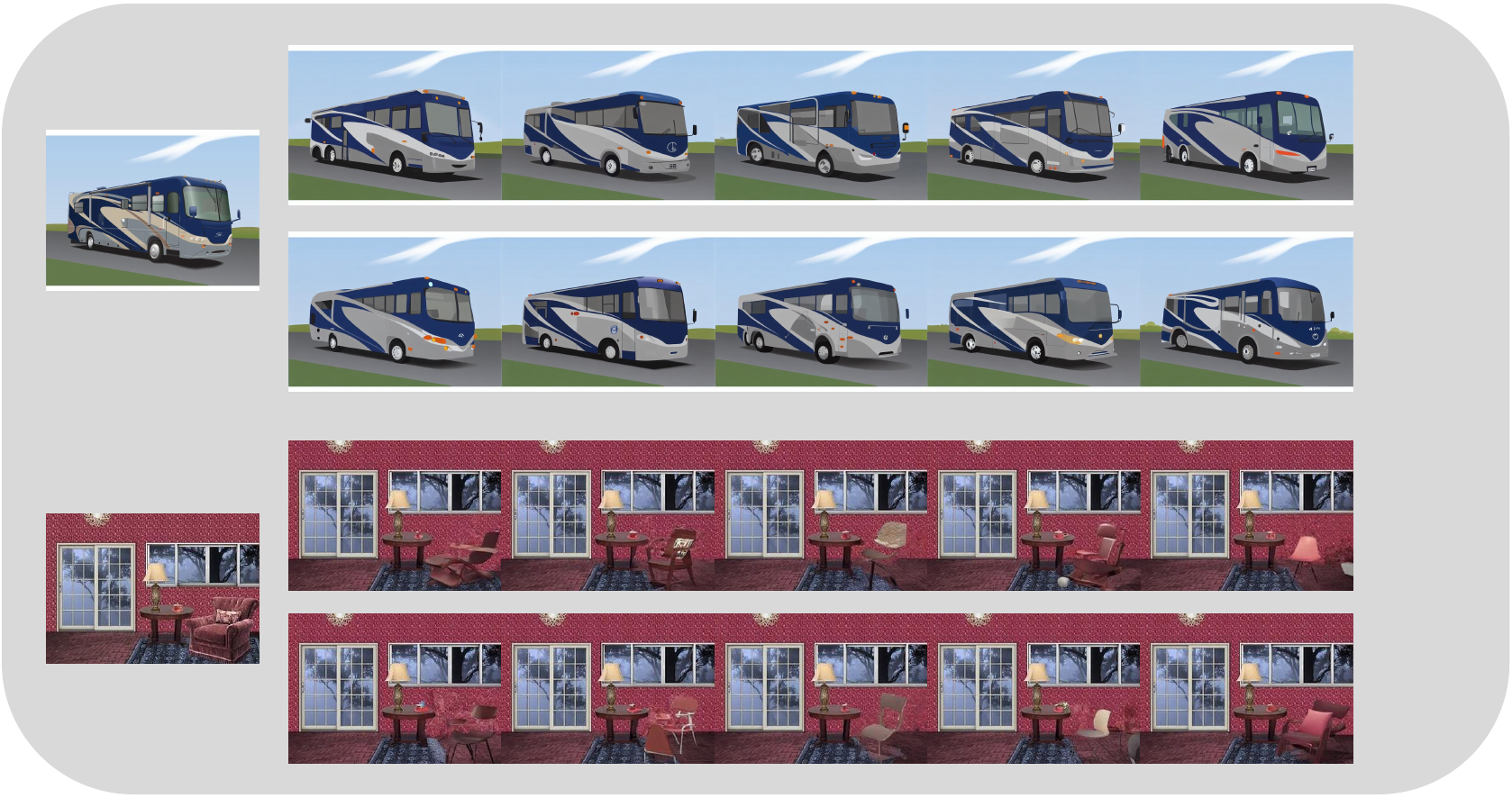}
        \caption{(left) The original image of the Clipart1k. (right) The generated image of Clipart1k.}
        \label{fig:clip}
    \end{minipage}
\end{figure}
\textbf{Clipart1k}\cite{inoue2018cross}: A stylized dataset featuring cartoon-like illustrations with coarse-grained categories (e.g., common indoor objects), optimized for cross-domain few-shot object detection (CDFSOD) due to its synthetic visual simplicity and domain-shift challenges.\\

\textbf{DIOR}\cite{li2020object}: A large-scale remote sensing dataset for aerial object detection, containing annotations for 20 categories, including airplanes, ships, golf courses, and other geospatial objects captured in satellite imagery.\\
\begin{figure}[htbp]
    \centering
    \begin{minipage}[b]{0.48\textwidth}  
        \centering
        \includegraphics[width=\linewidth]{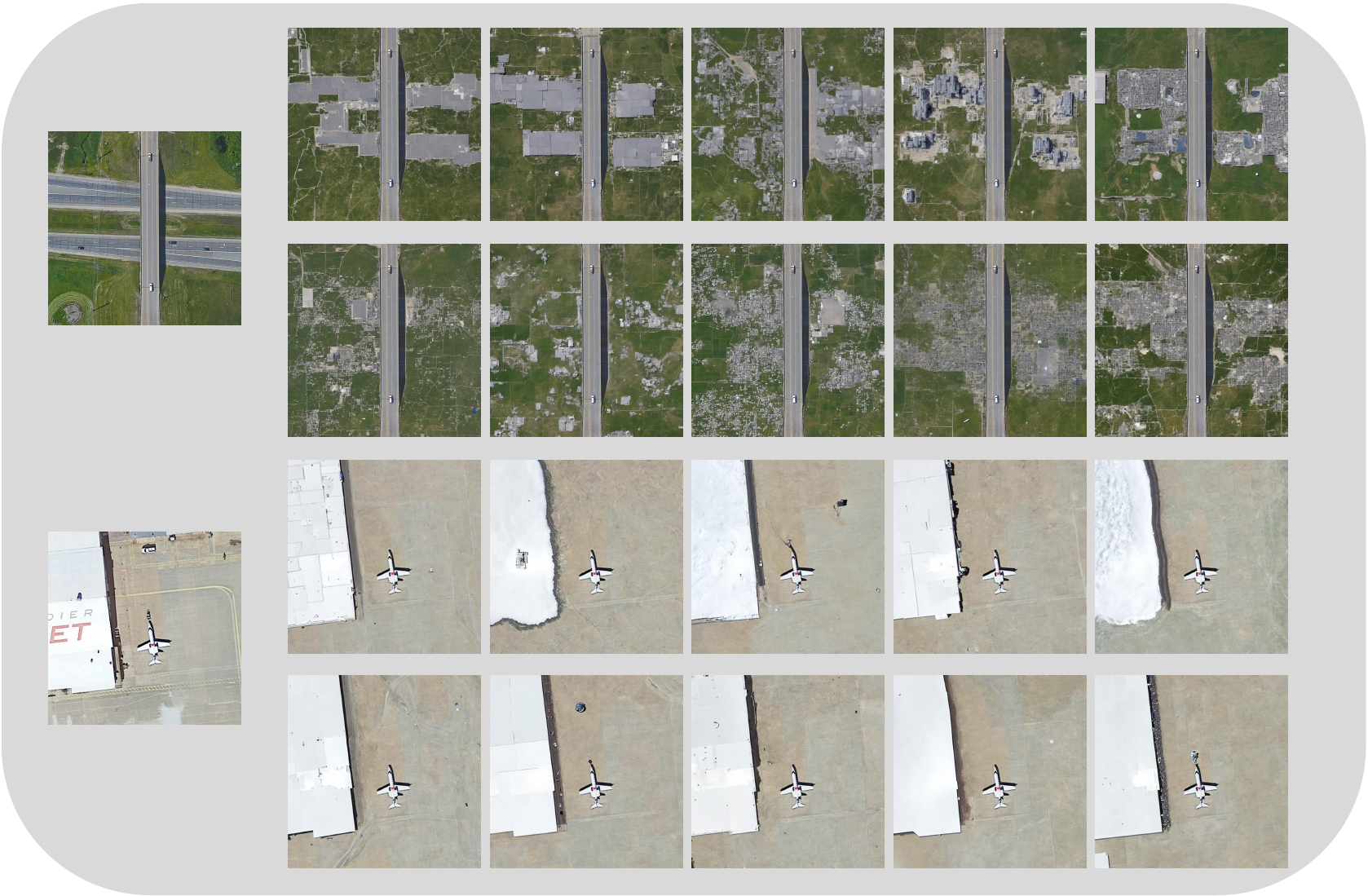}
        \caption{(left) The original image of the DIOR. (right) The generated image of DIOR.}
        \label{fig:dior}
    \end{minipage}
    \hfill
    \begin{minipage}[b]{0.48\textwidth}  
        \centering
        \includegraphics[width=\linewidth]{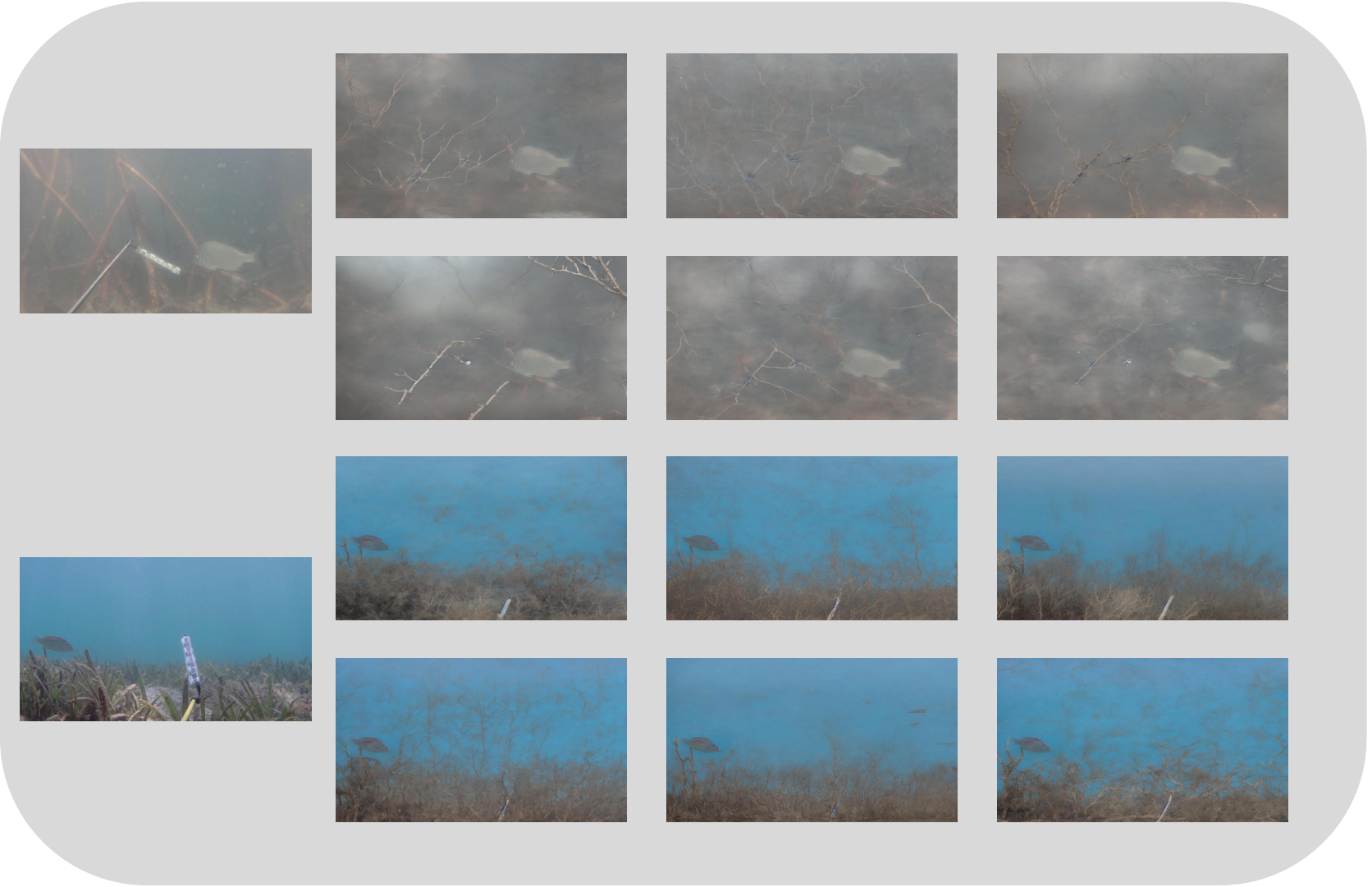}
        \caption{(left) The original image of the DeepFish. (right) The generated image of DeepFish.}
        \label{fig:fish}
    \end{minipage}
\end{figure}

\textbf{DeepFish}\cite{saleh2020realistic}: A high-resolution underwater image dataset capturing marine biodiversity across varied natural habitats, tailored for ecological monitoring and species identification in complex aquatic environments.

\textbf{NEU-DET}\cite{song2013noise}: An industrial inspection dataset compiling six types of hot-rolled steel strip surface defects (e.g.,rolled-in scale, crazing, pitted surfaces, patches, inclusions, and scratches), critical for automated quality control systems.

\begin{figure}[htbp]
    \centering
    \begin{minipage}[b]{0.48\textwidth}
        \centering
        \includegraphics[width=0.8\linewidth]{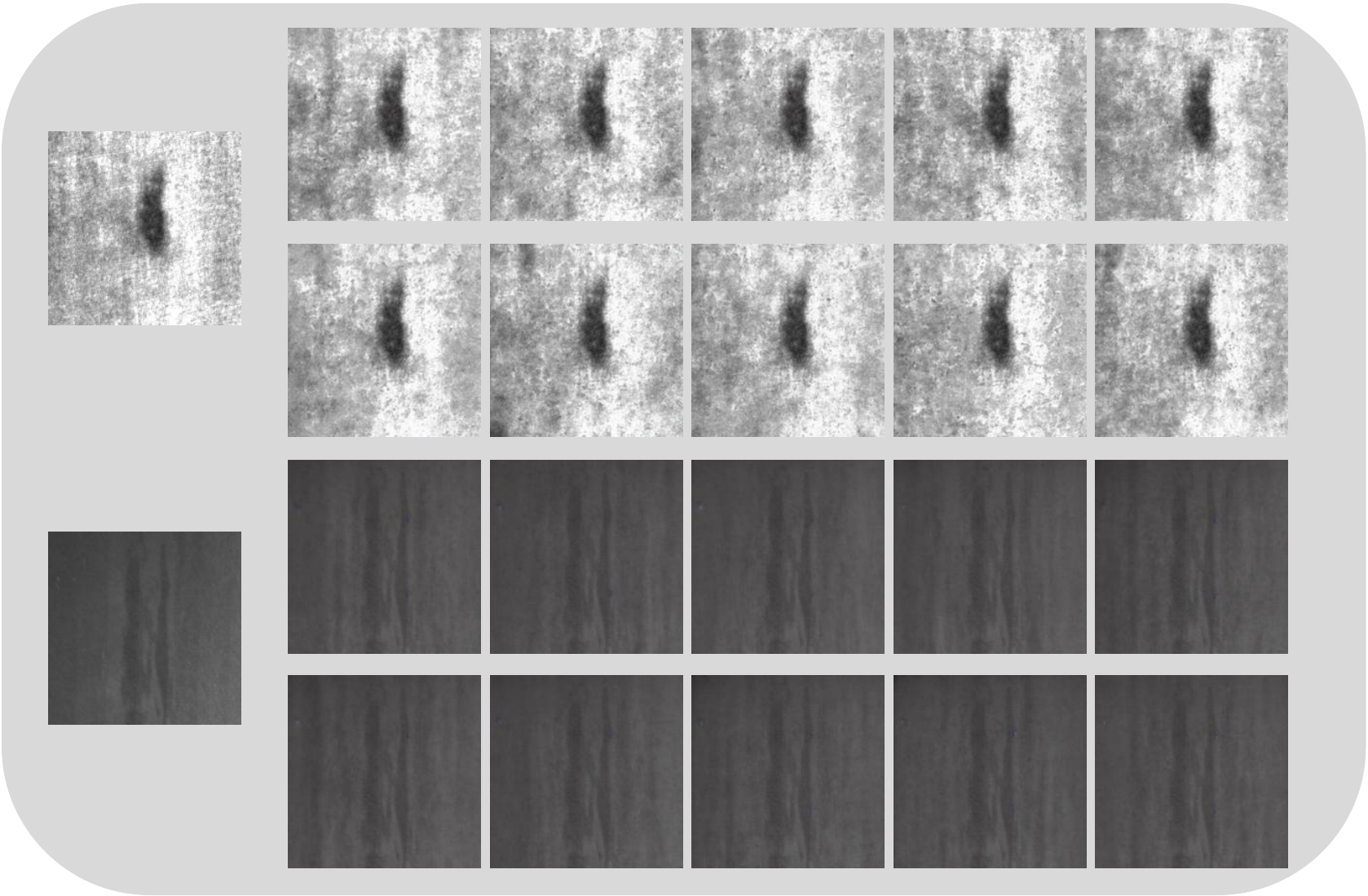}
        \caption{(left) The original image of the NEU-DET. (right) The generated image of NEU-DET.}
        \label{fig:neu-det}
    \end{minipage}
    \hfill
    \begin{minipage}[b]{0.48\textwidth}
        \centering
        \includegraphics[width=1\linewidth]{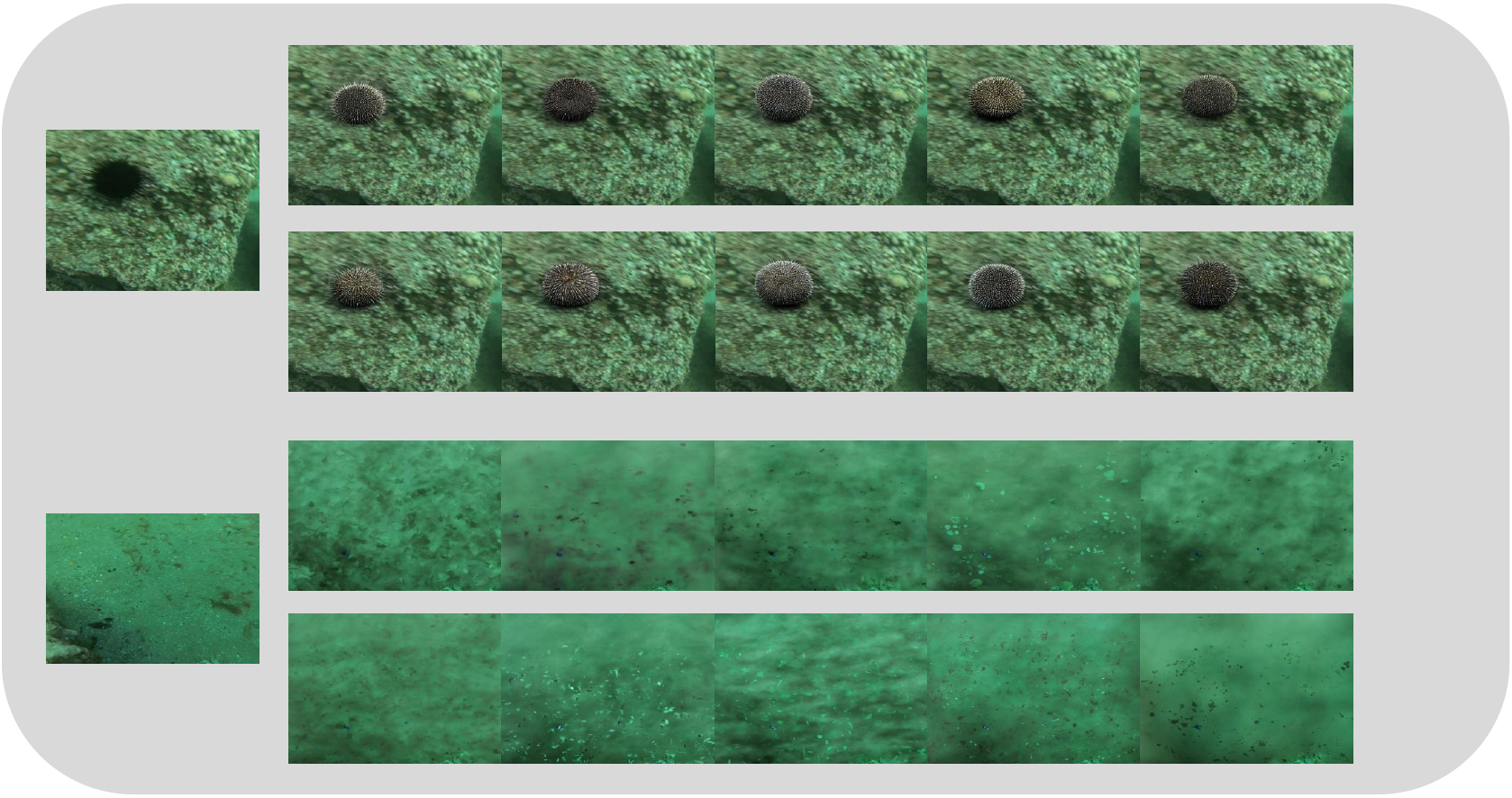}
        \caption{(left) The original image of the UODD. (right) The generated image of UODD.}
        \label{fig:uodd}
    \end{minipage}
\end{figure}

\textbf{UODD}\cite{jiang2021underwater}: An underwater object detection dataset featuring benthic organisms such as sea urchins, sea cucumbers, and scallops, addressing challenges in marine robotics and habitat analysis under low-visibility conditions.


\section{Implementation Details}
\subsection{Object Detection Model}
The learning rates of each dataset are shown in Tab.~\ref{tab: Learning rate}. The training loss, training epochs, and other hyperparameters in this study strictly follow the experimental setup in ~\cite{fu2024cross}. The loss function $L$ adopts the five-component structure, including $L_{loc}$, $L_{cls}$, $L_{dp}$ (consist of $L_{domain}$, $L_{proto}$, $L_{proto_{cls}}$), and its theoretical foundation is elaborated in the Methods section. 

\begin{equation}
L=L_{loc} + L_{cls} , L_{dp}
\end{equation}
\begin{equation}
L_{dp} = L_{domain}+ L_{proto} + L_{proto_{cls}}
\end{equation}

\begin{table}[htpb]
  \centering
  \caption{Learning rate of 1-shot, 5-shot, 10-shot settings}
  \vspace{-0.3cm}

  \resizebox{0.6\textwidth}{!}{
    \begin{tabular}{ccccccc}
    \hline
    Setting & Arta.& Clip.& DIOR  & Deep.& NEU.& UODD \\
    \hline
    1-shot & 0.0025 & 0.002 & 0.001 & 0.001 & 0.001 & 0.0025 \\
    5-shot & 0.0015 & 0.001 & 0.002 & 0.00075 & 0.0025 & 0.0025 \\
    10shot & 0.00125 & 0.00075 & 0.0015 & 0.00075 & 0.002 & 0.0025 \\
    \hline
    \end{tabular}%
    }
  \vspace{-0.6cm}
    
  \label{tab: Learning rate}%
\end{table}%

\subsection{Diffusion Model}
We use the Qwen3 to generate general prompts based on 10-shot samples (which include both 5-shot and 1-shot) for each dataset.

We employ the Stable-Diffusion v1.5 model. The sampling step range is set between 5 and 50, dynamically controlled by the strength coefficient (strength $\in [0.1, 1.0]$).

\begin{table}[h]
  \centering
  \caption{Top-k of each dataset.}
  \vspace{-0.3cm}
  
  \resizebox{0.6\textwidth}{!}{
    \begin{tabular}{ccccccc}
    \hline
    Value & ArTa. & Clip. & DIOR  & Deep. & NEU. & UODD \\
    \hline
    k     & 7     & 8     & 3     & 5     & 10    & 3 \\
    \hline
    \end{tabular}%
    }
  \vspace{-0.6cm}
    
  \label{tab: topk}%
\end{table}%




\section{Top-$k$ in Selection Module}
\subsection{$k$ of each dataset}
In most datasets, the model's performance does not necessarily improve as the number of generated images increases; this may be attributed to overfitting induced by a specific generated image. However, the generated images are generally effective, as models trained with them consistently outperform the baseline in overall performance.

We employ the top-k selection algorithm to filter the generated images, with the value of k tailored to each target domain dataset (Tab.~\ref{tab: topk}).

\subsection{Sensitivity of $k$}

The average performance of fixing $k$ vs. tailored $k$ for each dataset is 29.3\% vs. 32.1\% under the 1-shot setting, showing tailoring $k$ slightly improves the performance.
To verify the robustness of the choice of $k$, we plot the sensitivity analysis of $k$ on Artaxor, where the performance is stable between 5 and 10, showing the robustness.
\begin{figure}[htbp]
    \centering
    \vspace{-0.4cm}
    \includegraphics[width=0.7\linewidth]{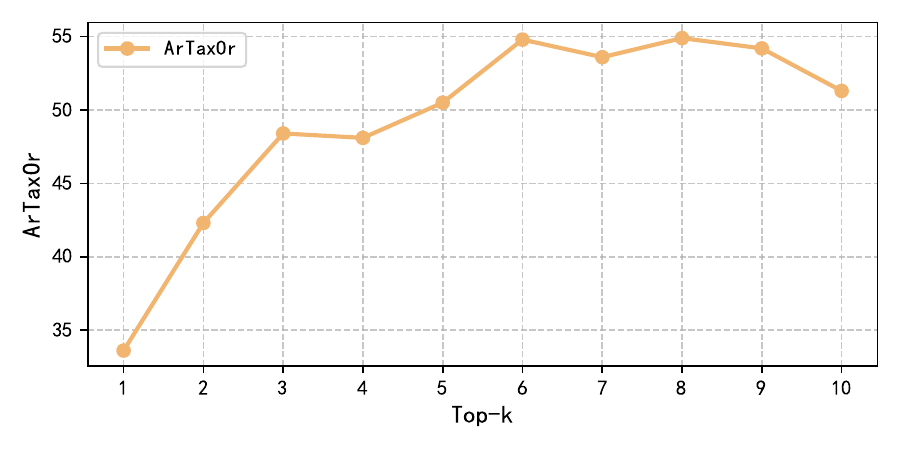}
    \vspace{-0.3cm}
    \caption{Ablation experiment of Top-k based on ETS.}    \vspace{-0.6cm}
    \label{fig: k-value}
\end{figure}

\section{More Backbone of our method}

To comprehensively evaluate the impact of backbone architecture on CDFSOD performance, we conduct a comparative analysis between the Vision Transformer Base (VIT-B) and the Vision Transformer Small (VIT-S) under the 1-shot setting. The experimental results are summarized in Tab.~\ref{tab: The performance of VITB and VITS}.

\begin{table*}[htbp]
  \centering
  \caption{The performance of VITB and VITS. $\dagger$ means we produce results. Avg. means average result.}
    \begin{tabular}{cccccccc}
    \hline
    Model & ArTaxOr & Clipart1k & DIOR  & DeepFish & NEU-DET & UODD  & Avg.\\
    \hline
    VITL  & 21.1  & 19.1  & 17.9  & 19.9  & 4.2   & 3.1   & 14.2 \\
    \cellcolor{cyan!10}VITL+\textbf{Ours}$\dagger$ & \cellcolor{cyan!10}26.7  & \cellcolor{cyan!10}22.3  & \cellcolor{cyan!10}19.8  & \cellcolor{cyan!10}22.1  & \cellcolor{cyan!10}5.4   & \cellcolor{cyan!10}5.6   & \cellcolor{cyan!10}16.9 \\
    VITB  & 20.5  & 17.8  & 17.2  & 20.2  & 3.4   & 3.17  & 13.7 \\
    \cellcolor{cyan!10}VITB+\textbf{Ours}$\dagger$ & \cellcolor{cyan!10}25.8  & \cellcolor{cyan!10}20.9  & \cellcolor{cyan!10}19.6  & \cellcolor{cyan!10}21.9  & \cellcolor{cyan!10}4.9   & \cellcolor{cyan!10}4.7   & \cellcolor{cyan!10}16.3 \\
    VITs  & 21.1  & 18    & 17.2  & 19.9  & 3.5   & 3.2   & 13.8 \\
    \cellcolor{cyan!10}VITS+\textbf{Ours}$\dagger$ & \cellcolor{cyan!10}26.1  & \cellcolor{cyan!10}21.1  & \cellcolor{cyan!10}19.2  & \cellcolor{cyan!10}20.9  & \cellcolor{cyan!10}4.7   & \cellcolor{cyan!10}4.5   & \cellcolor{cyan!10}16.1 \\
    \hline
    \end{tabular}%
  \label{tab: The performance of VITB and VITS}%
\end{table*}%


\section{Prompts of Foreground Inpaint}
\textbf{Artaxor}\\  
\noindent{\color{blue}\textit{\noindent{\color{blue}\textit{$prompt$:}}}}"Ultra HD macro shot of a \{category name\} specimen, dorsal view displaying hardened elytra with parallel striations, segmented antennae, and tarsal claws clearly visible.Clinical lighting, scanning electron microscope level detail"  \\
\noindent{\color{blue}\textit{\noindent{\color{blue}\textit{$negative\ prompt$:}}}} "background, plants, ground, sky, shadow, human, man-made, blur, text, low-res, deformed, extra limbs, merged specimens, unnatural lighting."

\textbf{Clipart1k}\\ 
\noindent{\color{blue}\textit{$prompt$:}}"A simple cartoon-style illustration of \{category name\}. Using a color palette. Clean lines, flat design, no shadows, a white/plain background, and a minimalist composition. "  \\
\noindent{\color{blue}\textit{$negative\ prompt$:}} "low quality, blurry, watermark, text, signature, logo, extra limbs, deformed hands, distorted face, dark shadows, noise, grain, over-saturated, dull colors, messy lines, abstract art, multiple subjects."

\textbf{DIOR}\\ 
\noindent{\color{blue}\textit{$prompt$:}}"High-resolution satellite view of \{category name\}, ultra-realistic, 8K detail, professional photography, drone perspective, God's eye view."  \\
\noindent{\color{blue}\textit{$negative\ prompt$:}} "blurry, low resolution, cartoonish, distorted perspective, close-up view, heavy shadows, over-saturated colors, unrealistic lighting, human figures, text overlays, digital artifacts."

\textbf{FISH}\\
\noindent{\color{blue}\textit{$prompt$:}}"natural camouflage, A \{category name\} is swimming. (brown decaying branches:1.1), murky water with particles, dim lighting. "  \\
\noindent{\color{blue}\textit{$negative\ prompt$:}} "clear water, tropical fish, bright colors, sharp details, cartoonish style, unrealistic lighting."

\textbf{NEU-DET}\\
\noindent{\color{blue}\textit{$prompt$:}}"Abstract grayscale texture: \{category name\}, old plaster wall, high-contrast blotches, ultra-low resolution, film grain effect, no recognizable objects, microscopic view, RAW unprocessed"  \\
\noindent{\color{blue}\textit{$negative\ prompt$:}} "sharp focus, high detail, vibrant colors, human figures, landscapes, modern art, clean lines, symmetry, digital art, 3D rendering, cartoon, anime, text, logos, recognizable shapes, faces, architecture, nature."  
\\

\textbf{UODD}\\
\noindent{\color{blue}\textit{$prompt$:}}"Ultra-detailed close-up of a \{category name\} in isolation, showcasing"  \\
\noindent{\color{blue}\textit{$negative\ prompt$:}} "clear water, high visibility, bright lighting, vibrant colors, humans, ships, coral reefs, tropical fish, sharp edges, overexposed, cartoonish, anime, text, logos, artificial objects."

\section{Prompts of Foreground Inpaint}
\textbf{Artaxor}\\  
\noindent{\color{blue}\textit{$prompt$:}}"Ultra HD macro shot of a \{category name\} specimen, dorsal view displaying hardened elytra with parallel striations, segmented antennae, and tarsal claws clearly visible.Clinical lighting, scanning electron microscope level detail"  \\
\noindent{\color{blue}\textit{$negative\ prompt$:}} "background, plants, ground, sky, shadow, human, man-made, blur, text, low-res, deformed, extra limbs, merged specimens, unnatural lighting"  

\textbf{Clipart1k}\\ 
\noindent{\color{blue}\textit{$prompt$:}}"A simple cartoon-style illustration of \{category name\}. Using a color palette. Clean lines, flat design, no shadows, a white/plain background, and a minimalist composition. "  \\
\noindent{\color{blue}\textit{$negative\ prompt$:}} "low quality, blurry, watermark, text, signature, logo, extra limbs, deformed hands, distorted face, dark shadows, noise, grain, over-saturated, dull colors, messy lines, abstract art, multiple subjects"  

\textbf{DIOR}\\ 
\noindent{\color{blue}\textit{$prompt$:}}"High-resolution satellite view of \{category name\}, ultra-realistic, 8K detail, professional photography, drone perspective, God's eye view"  \\
\noindent{\color{blue}\textit{$negative\ prompt$:}} "blurry, low resolution, cartoonish, distorted perspective, close-up view, heavy shadows, over-saturated colors, unrealistic lighting, human figures, text overlays, digital artifacts"

\textbf{FISH}\\
\noindent{\color{blue}\textit{$prompt$:}}"underwater scene, natural camouflage, A \{category name\} is swimming. (brown decaying branches:1.1), murky water with particles, dim lighting "  \\
\noindent{\color{blue}\textit{$negative\ prompt$:}} "clear water, tropical fish, bright colors, sharp details, cartoonish style, unrealistic lighting"  

\textbf{NEU-DET}\\
\noindent{\color{blue}\textit{$prompt$:}}"Abstract grayscale texture: \{category name\}, old plaster wall, high-contrast blotches, ultra-low resolution, film grain effect, no recognizable objects, microscopic view, RAW unprocessed"  \\
\noindent{\color{blue}\textit{$negative\ prompt$:}} "sharp focus, high detail, vibrant colors, human figures, landscapes, modern art, clean lines, symmetry, digital art, 3D rendering, cartoon, anime, text, logos, recognizable shapes, faces, architecture, nature"  
\\
\textbf{UODD}\\
\noindent{\color{blue}\textit{$prompt$:}}"Ultra-detailed close-up of a \{category name\} in isolation, showcasing"  \\
\noindent{\color{blue}\textit{$negative\ prompt$:}} "clear water, high visibility, bright lighting, vibrant colors, humans, ships, coral reefs, tropical fish, sharp edges, overexposed, cartoonish, anime, text, logos, artificial objects."  

\section{Prompts of Background Inpaint}
\textbf{Artaxor}\\
\noindent{\color{blue}\textit{$prompt$:}}"A macro photography, lying on a leaf/collecting nectar/hanging on a web" 
\noindent{\color{blue}\textit{$negative\ prompt$:}} "insect, blurry, low resolution, cartoonish, artificial lighting, heavy shadows, human figures, text overlay, unnatural colors, cluttered background, overexposed, underexposed, symmetry"

\textbf{Clipart1k}\\
\noindent{\color{blue}\textit{$prompt$:}}"A simple cartoon-style illustration. Using a color palette. Clean lines, flat design, no shadows, minimalistic composition. " \\
\noindent{\color{blue}\textit{$negative\ prompt$:}} "\{category name\}, low quality, blurry, watermark, text, signature, logo, extra limbs, deformed hands, distorted face, dark shadows, noise, grain, over-saturated, dull colors, messy lines, abstract art, multiple subjects"

\textbf{DIOR}\\
\noindent{\color{blue}\textit{$prompt$:}}"High-resolution satellite view, ultra-realistic, 8K detail, professional photography, drone perspective, God's eye view" \\
\noindent{\color{blue}\textit{$negative\ prompt$:}} "\{category name\}, blurry, low resolution, cartoonish, distorted perspective, close-up view, heavy shadows, over-saturated colors, unrealistic lighting, human figures, text overlays, digital artifacts"

\textbf{FISH}\\
\noindent{\color{blue}\textit{$prompt$:}}"underwater scene, natural camouflage, (brown decaying branches:1.1), murky water with particles, dim lighting" \\
\noindent{\color{blue}\textit{$negative\ prompt$:}} "\{category name\}, clear water, tropical fish, bright colors, sharp details, cartoonish style, unrealistic lighting"

\textbf{NEU-DET}\\
\noindent{\color{blue}\textit{$prompt$:}}"Abstract grayscale texture: old plaster wall, high-contrast blotches, ultra-low resolution, film grain effect, no recognizable objects, microscopic view, RAW unprocessed" \\
\noindent{\color{blue}\textit{$negative\ prompt$:}} "sharp focus, high detail, vibrant colors, human figures, landscapes, modern art, clean lines, symmetry, digital art, 3D rendering, cartoon, anime, text, logos, recognizable shapes, faces, architecture, nature"

\textbf{UODD}\\
\noindent{\color{blue}\textit{$prompt$:}}"Underwater scene, dark green and teal water with low visibility, scattered pebbles/seaweed fragments, subtle organic stains or floating particles, dim and diffused sunlight from above, realistic photography style, muted color palette" \\
\noindent{\color{blue}\textit{$negative\ prompt$:}} "\{category name\}, clear water, high visibility, bright lighting, vibrant colors, humans, ships, coral reefs, tropical fish, sharp edges, overexposed, cartoonish, anime, text, logos, artificial objects."


%
%
\bibliographystyle{splncs04}
\bibliography{main}